\documentclass{article} 
\usepackage{iclr2027_conference,times}

\usepackage{amsmath,amsfonts,bm}

\def\eqref#1{equation~\ref{#1}}

\def\1{\bm{1}}

\DeclareMathAlphabet{\mathsfit}{\encodingdefault}{\sfdefault}{m}{sl}
\SetMathAlphabet{\mathsfit}{bold}{\encodingdefault}{\sfdefault}{bx}{n}

\usepackage[utf8]{inputenc} 
\usepackage[T1]{fontenc}    
\usepackage{hyperref}       
\usepackage{url}            
\usepackage{amsfonts}       
\usepackage{nicefrac}       
\usepackage{enumitem}
\usepackage{microtype}      
\usepackage{xcolor}         

\usepackage{booktabs} 
\usepackage{makecell}
\usepackage{textcomp}
\usepackage{amsmath, amsthm}

\usepackage[ruled, vlined, linesnumbered]{algorithm2e}
\usepackage[noend]{algpseudocode}
\usepackage{pdfpages}
\usepackage{graphicx}
\usepackage{amssymb}
\usepackage{mathrsfs}
\usepackage{epsfig}
\usepackage{graphicx}
\usepackage{multirow}
\usepackage{balance}
\usepackage{wrapfig}
\usepackage[normalem]{ulem}
\usepackage{threeparttable}
\usepackage{array}
\newcolumntype{C}[1]{>{\centering\arraybackslash}p{#1}}
\usepackage{lineno}

\usepackage{bbm}

\usepackage{graphicx}
\usepackage{float}
\usepackage{subfig}

\usepackage{pifont}

\title{Goal-Conditioned Supervised Learning for LLM Fine-Tuning}

\author{
   Shijun Li$^{1}$,
   Kaiwen Dong$^{2}$,
   Xiang Gao$^{2}$, 
   Joydeep Ghosh$^{4}$ 
   \\
  $^{1,4}$The University of Texas at Austin, $^{2}$Intuit AI Research \\
  \texttt{shijunli@utexas.edu}\\
  \ \ \
  }

\iclrfinalcopy 
\begin{document}

\maketitle
\thispagestyle{fancy}
\fancyhead[L]{Preprint. Under review.} 

\begin{abstract}
Large language models often require fine-tuning to better align their behavior with user intent at deployment. Existing approaches are commonly divided into online and offline paradigms. Online methods, such as RL-based alignment, can directly optimize outcome quality but typically rely on external reward models and iterative rollouts, making them costly and difficult to deploy in many cases. Offline methods are more efficient, but prevailing approaches such as supervised fine-tuning (SFT) and direct preference optimization (DPO) remain limited: SFT typically collapses graded feedback into binary supervision, while DPO depends on paired preference data that is often unavailable or expensive to construct.
In this paper, we propose goal-conditioned supervised learning (GCSL) as an offline fine-tuning framework for LLMs. Our core idea is to treat feedback signals directly as an explicit goal and train the model, purely through supervised learning, to generate responses that achieve that goal.
To better exploit graded feedback, we further introduce a novel goal formulation that defines learning as consistently pursuing outcomes above a target quality threshold, rather than imitating samples from a selected high-quality subset. 
This design mitigates the bounded-learning effect by learning transferable patterns of meeting or exceeding quality thresholds. 
We also propose natural-language goal representations to further connect these patterns to the LLM’s pretrained knowledge and generalization capabilities.
We evaluate our method on three tasks: non-toxic generation, code generation, and LLM for recommendation. Results show that our approach consistently outperforms standard offline fine-tuning baselines while retaining the efficiency, scalability, and simple data requirements of supervised learning. Code is available at: \href{https://anonymous.4open.science/r/GCSL_LLM-F5D5}{https://anonymous.4open.science/r/GCSL\_LLM-F5D5}.
\end{abstract}

\section{Introduction}\label{sec:intro}

LLMs exhibit strong general capabilities, but their pretraining objective often doesn't reliably produce behaviors users want at deployment. Consequently, fine-tuning has become the standard route to alignment. Existing fine-tuning methods can be broadly categorized into online and offline paradigms. Online approaches, most notably RL-based alignment such as PPO/GRPO \citep{grpo}, optimize model behavior through iterative sampling and updates driven by a reward signal. While these methods can deliver strong performance by directly optimizing an outcome objective, they may come with substantial practical limitations: they typically rely on an external reward model, which is expensive to train and can be mis-specified relative to real user needs, thereby introducing harmful noise and biases into optimization \citep{scheid2024optimal, gao2023scaling}. In addition, online rollouts and iterative updates are time- and resource-consuming, making such methods difficult to scale or deploy in many real-world settings \citep{rafailov2023direct, qiu2025evolution}.

Such constraints have motivated the widespread use of offline fine-tuning methods. Most current offline fine-tuning approaches can be broadly categorized as supervised fine-tuning (SFT) or direct preference optimization (DPO) (and closely related preference-based objectives) \citep{rafailov2023direct, meng2024simpo}. DPO avoids online RL by learning from preference comparisons, but it requires training data in a paired format (preferred vs. not preferred outputs), which is not always available or easy to construct. In contrast, SFT can directly leverage raw sequential data collected in the wild, such as user dialogue records, interaction logs in recommender systems, or other behavioral traces, making it broadly applicable and simple to operationalize. However, this flexibility comes with an important limitation: in practice, SFT often reduces available niche feedback or reward signals into a binary notion of correctness (typically via a handcrafted cutoff) \citep{dong2023raft}, and then imitates the selected “positive” subset. This thresholding discards fine-grained information contained in graded feedback, makes performance sensitive to the cutoff, and treats all positive samples as equally good demonstrations, which can bias learning towards the average quality of the selected subset rather than explicitly encouraging consistent improvement towards higher-quality outcomes.

This paper investigates goal-conditioned supervised learning (GCSL) \citep{TrajectoryTransformer, liu2022goal} as an offline fine-tuning paradigm for LLMs. Many real deployments naturally provide feedback signals such as numeric ratings or categorical judgments. Instead of converting these signals into homogeneous demonstrations (SFT), pairwise comparisons (DPO), or rewards requiring reward model and online RL (PPO/GRPO), GCSL suggests a direct reframing: treat feedback as an explicit goal and train the model via supervised learning to generate responses that achieve the goal.


This brings three key advantages. First, it enables direct use of feedback (scores or categories) without external reward models or paired samples. Second, it realizes long-horizon goal-achieving optimization within a pure supervised framework, thereby retaining the efficiency and scalability of standard supervised training and avoiding online rollouts and reward-model-dependent iteration. 
Third, and most importantly, our approach is designed to mitigate a key limitation of SFT and classic GCSL: learning is biased toward the average quality of the selected training subset. To address this, we introduce a novel goal-achieving objective that defines the goal as consistently pursuing outcomes above a given quality threshold. This formulation better exploits graded feedback, avoids collapsing supervision into undifferentiated positive samples or pairwise contrastive samples, and encourages the model to generalize goal-seeking behavior across diverse targets in the data. In other words, the new goal-achieving objective seeks to drive directional progression of performance toward higher-quality outcomes through goal-conditioned generalization.

The closest prior work in this direction is Quark \citep{quark}, which incorporates goal-conditioned ideas into LLM optimization but retains several drawbacks. First, Quark still relies on an online procedure with an external reward model, reducing efficiency and inheriting reward-model mis-specification risks. Second, its goal definition via the highest quantized score bin still collapses toward SFT-like behavior, limiting improvement beyond the average quality within that bin. Finally, its special-token goal representation underutilizes LLM’s native semantic understanding and extrapolation capability. These issues motivate our approach: a purely offline supervised formulation that directly leverages feedback as goals, reformulates goals to better reflect the optimization target, and uses natural-language goal representations to better exploit LLM's semantic ability and knowledge.

Our key contributions can be summarized as follows:

\begin{itemize}[leftmargin=*]

    \item We reframe LLM fine-tuning as goal-conditioned supervised learning, enabling training from direct feedback signals while avoiding online reward-model-dependent training and paired preference data, benefiting from high training efficiency, scalability, and data generality.

    \item We introduce a novel goal-achieving objective to mitigate an important imitation tendency of SFT and classic GCSL, whose learning 
     is biased toward the average quality
    of the selected training subset. By formulating learning as consistently pursuing outcomes above a target quality threshold, our approach better exploits graded feedback, mitigates bounded-learning effects, and explicitly drives directional progression of quality. We further propose natural-language goal representations, which better leverage the LLM’s inherent semantic understanding and extrapolation abilities.

    \item We evaluate our approach on non-toxic generation, code generation, and recommendation tasks to demonstrate its generality and effectiveness. Empirical results show that it consistently outperforms standard offline fine-tuning baselines while retaining high efficiency and simple data requirements.
    
\end{itemize}




\section{Related Work}\label{sec:related}

\noindent \textbf{LLM Fine-Tuning.}
Fine-tuning is the standard approach for aligning pretrained LLMs with user intent and deployment constraints. Supervised fine-tuning (SFT) is widely used for its simplicity and stability, but with non-binary feedback (e.g., scalar ratings), it typically relies on handcrafted cutoffs to select “good” samples and then treats all retained positives equally. This makes performance sensitive to the cutoff and biases learning toward the average quality of the selected subset rather than consistently improving toward higher-quality outcomes. To move beyond imitation, many pipelines adopt RL-based fine-tuning (e.g., PPO/GRPO), which usually requires an external reward model and iterative online updates. Such reward models are costly to train and often mis-specified, introducing noise and bias into optimization, while online rollouts add substantial computational cost and complexity. Preference-based methods such as DPO \citep{rafailov2023direct} avoid explicit RL, but they require paired preference data, which may be unavailable in many scenarios. Our work instead targets a more direct and efficient alternative: optimizing outcome quality using only supervised learning and readily available scalar or categorical feedback. See discussions of our approach with other methods such as KTO \citep{ethayarajh2024kto} and CA \citep{hu2023aligning} in Appendix \ref{sec:exp_rel}.


\noindent \textbf{Goal-Conditioned Supervised Learning.}
Goal-conditioned supervised learning (GCSL) offers an RL-like paradigm while remaining purely supervised. By conditioning a policy on an explicit goal and learning from supervised targets, GCSL can train goal-achieving behavior without value estimation or policy-gradient optimization. The definition of goals in GCSL can be broad, such as final states, cumulative rewards, or specific outcomes \citep{TrajectoryTransformer, liu2022goal}.


A closely related attempt to bring GCSL on LLMs is Quark \citep{quark}, which introduces goal-conditioned optimization but remains an online method requiring a reward model. More importantly, its goals are tied to specific quantized score bins, which can reduce goal-conditioning to imitation of these subsets and limit gains beyond the average quality. By contrast, our method reformulates goals as achieving outcomes above a target quality threshold, yielding a more aligned optimization objective for inference target within a purely offline supervised learning framework. We further express goals in natural language to better leverage LLM’s semantic understanding and reasoning abilities. SteerLM \citep{dong2023steerlm} is similar to Quark, still requiring the reward model for online update.
Some recent work also draws on GCSL ideas for LLM-related learning, but their formulations differ substantially from ours. E.g., \cite{nath2024learning} use GCSL-inspired ideas to train a reward model conditioned on future goals, whereas PNLC~\citep{hong2025planning} learns a Q-function to evaluate actions for reaching a target goal state. Since these methods are not designed to directly apply GCSL to LLM fine-tuning, we don't include them in our comparisons. We also include discussion of similar ideas explored in our work like learning from graded reward signals with previous research in fields beyond LLM like Upside-Down RL \citep{schmidhuber2019reinforcement} in Appendix \ref{sec:exp_rel}.

\section{Methodology}

In this section, we first formulate LLM fine-tuning with offline feedback as a goal-conditioned sequence modeling problem. We then describe a classic offline adaptation of GCSL for LLM fine-tuning, analyze its limitations, and finally introduce our beyond-threshold formulation, \textbf{GCSL-bey}, together with its natural-language variant, \textbf{GCSL-bey-NL}.

\subsection{Classic GCSL for LLM Fine-Tuning}\label{sec:classic_gcsl}

\paragraph{Problem setting.}
We view an autoregressive LLM as a goal-conditioned policy. Given an input prompt and/or context $x$, the model generates a response sequence $s=(y_1,\ldots,y_T)$ token by token, where each next-token decision can be viewed as an action and the full response as a trajectory toward some target outcome. Our offline training set is: $\mathcal{D}=\{(x_i, s_i, r_i)\}_{i=1}^N,$
where $r_i$ denotes the feedback signal associated with the complete response $y_i$. Depending on the application, $r_i$ can be a scalar score (e.g., non-toxicity level, code efficiency score) or an ordered categorical judgment (e.g., user's categorical rating). Importantly, unlike online RL-based alignment methods, we assume that these feedback signals are already available in the logged data, either from direct user feedback or from pre-existing task evaluators. Therefore, no online rollout, reward-model fitting, or iterative re-scoring is required during the offline fine-tuning.

\paragraph{Offline reward quantization.}
Following the standard approach of classic GCSL research like Trajectory Transformer \citep{TrajectoryTransformer} and Quark \citep{quark}, we first convert the feedback signals in the training data into a finite set of goal labels using equal-frequency binning. For scalar or ordered feedback, let $\tau_1 < \tau_2 < \cdots < \tau_K$
be ordered bin lower boundaries and let $Q(r)\in\{1,\ldots,K\}$ be the corresponding quantizer. Each quantized level is represented by a special goal token $[R_k]$. For each example $(x_i,s_i,r_i)$, we assign
\[
q_i = Q(r_i), \qquad g_i^{\mathrm{cls}} = [R_{q_i}],
\]
and construct the quantized offline dataset
\[
\widetilde{\mathcal{D}}_{\mathrm{cls}} = \{(x_i, s_i, g_i^{\mathrm{cls}})\}_{i=1}^N.
\]
If the feedback is already categorical, the category itself can be used directly as the goal label. Notably, unlike Quark, which repeatedly explores, re-scores, and re-quantizes newly sampled outputs, we perform this quantization once on a fixed offline dataset before fine-tuning.

\begin{figure}[t]
\centering
\begin{minipage}{0.85\linewidth}
    \centering
    \includegraphics[width=0.9\linewidth]{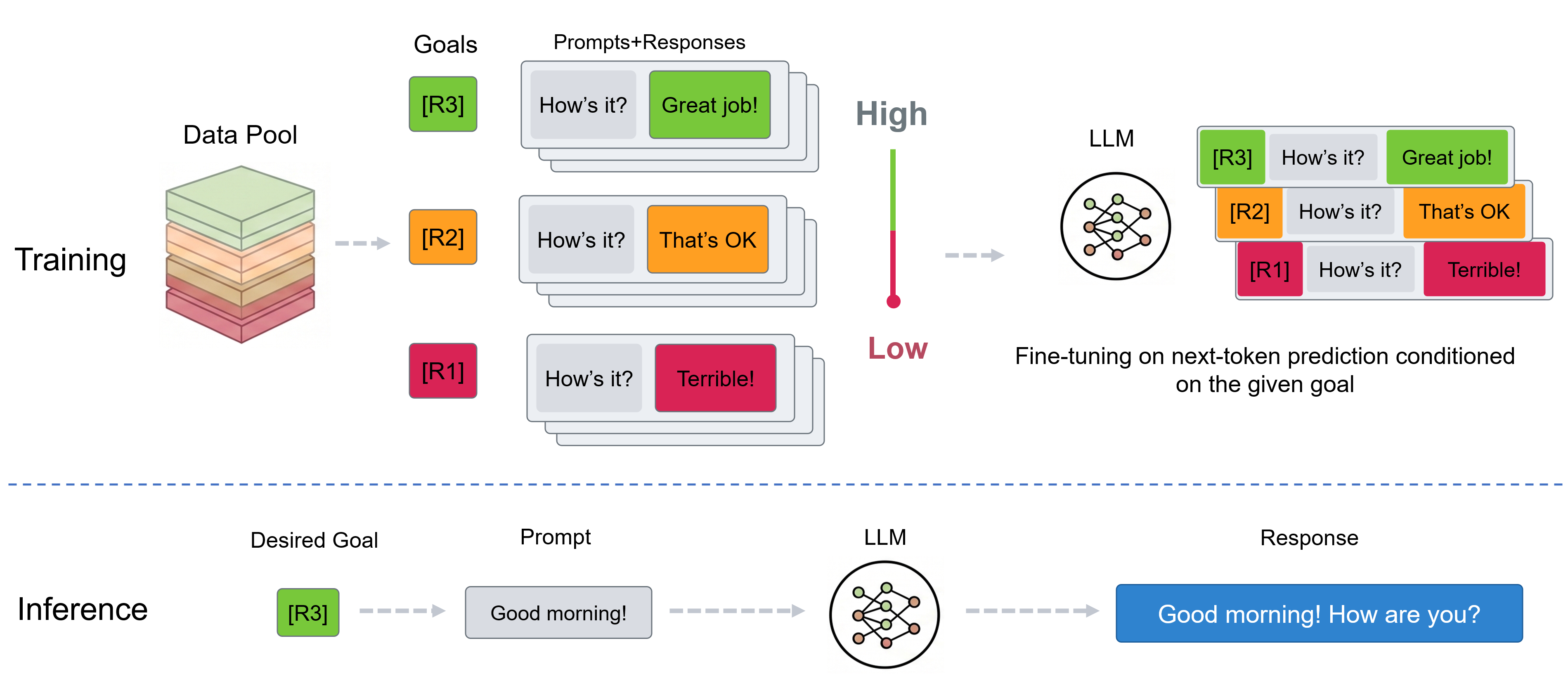}
\end{minipage}
\caption{Workflow of applying classic GCSL for LLM fine-tuning.}
\vspace{-2mm}
\label{fig:classic_gcsl}
\end{figure}

\paragraph{Goal-conditioned supervised fine-tuning.}
Given the quantized dataset, we fine-tune the LLM with standard teacher forcing conditioned on the goal:
\begin{equation}
\mathcal{L}_{\mathrm{cls}}(\theta)
=
-\sum_{i=1}^{N}\sum_{t=1}^{T_i}
\log p_\theta\!\left(y_{i,t}\mid x_i, g_i^{\mathrm{cls}}, y_{i,<t}\right).
\end{equation}
This is standard next-token prediction, except that the target response is conditioned not only on the input $x_i$ but also on the desired goal label. At inference time, given a new prompt $x$, we specify a goal token $g$ and decode from $p_\theta(\cdot\mid x,g)$. When the objective is to maximize response quality associated with the goal, the natural choice is the highest goal token $[R_K]$. Figure \ref{fig:classic_gcsl} illustrates the workflow of classic GCSL on fine-tuning LLM for non-toxicity generation.


\paragraph{Benefits of classic offline GCSL.}
Classic offline GCSL provides several practical advantages. First, it is a pure offline supervised method and therefore avoids the computational overhead of online rollouts, iterative policy updates, and reward model requirement. Second, it can directly use conventional sequential data with one-sided feedback signals, avoiding the need for paired preference data as in DPO. Third, compared with SFT, it makes finer-grained use of the training set by distinguishing samples with different quality levels rather than collapsing them into a binary positive/negative split. This helps the model learn general goal-achieving capabilities from diverse goal-reaching training patterns and adapt them for goal-achieving at inference.

Nevertheless, classic GCSL still exhibits two important limitations when applied to LLM fine-tuning.

\subsection{Deficiencies of Classic GCSL}\label{sec:def_GCSL}

\paragraph{SFT-like performance bottleneck.}
Classic GCSL defines goals as belonging to a quantized interval. In practice, when we want the best possible response, we condition on the highest goal token $[R_K]$ at inference time. However, the training examples associated with each goal token are simply the offline responses that fall inside the corresponding bin. As a result, the model is still encouraged to imitate the average behavior of the subset, rather than consistently pursuing possibly higher quality, which should instead be the learning objective that directly aligns with the inference target.
Thus, although classic GCSL goes beyond the binary splitting of SFT, its learning can remain biased toward average quality of the selected top subset, an imitation tendency we refer to as the bounded-learning effect.



\paragraph{Goal representation bottleneck.}
Classic GCSL represents goals with abstract reward tokens or score embeddings. These can encode relative ordering, but they don't explicitly tell what the goal means or what type of behavior is expected to achieve it. For LLMs whose pretrained capabilities are tightly coupled with natural-language understanding and instruction following, such symbolic goal representations can underutilize the model's semantic knowledge and extrapolation ability.

\subsection{Going Beyond Classic GCSL}\label{sec:bey_gcsl}

We address these issues from two directions: a new goal definition and a new goal representation.


\begin{figure}[t]
\centering
\begin{minipage}{0.98\linewidth}
    \centering
    \includegraphics[width=0.95\linewidth]{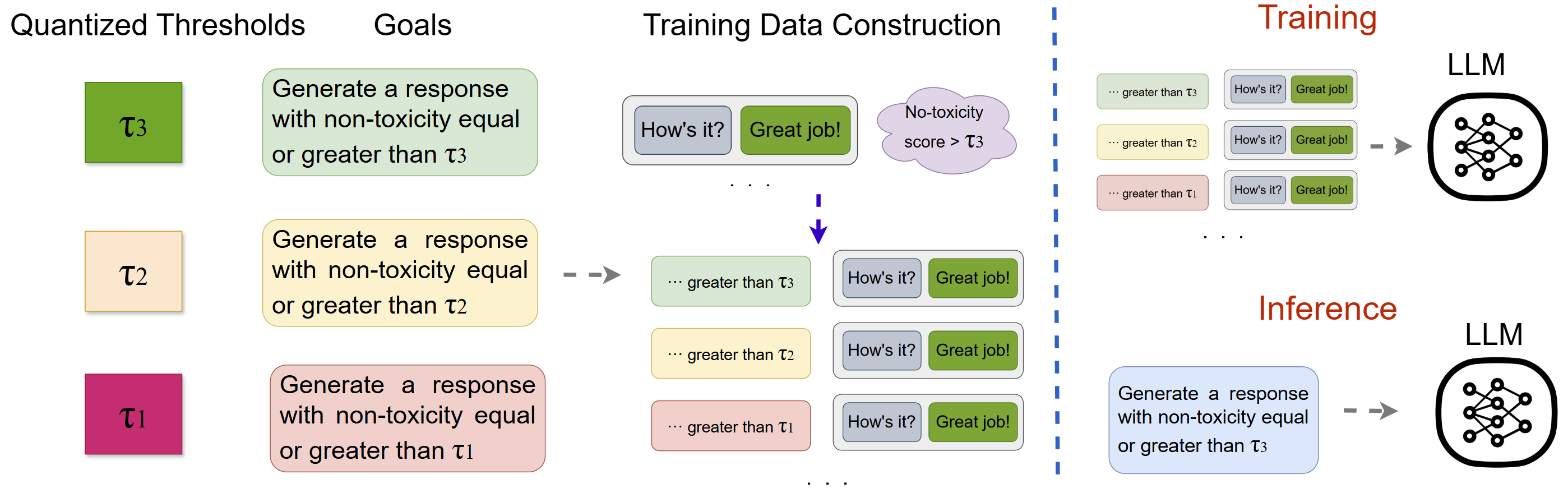}
\end{minipage}
\caption{Workflow of GCSL-bey-NL for LLM fine-tuning.}
\vspace{-2mm}
\label{fig:gcsl_bey}
\end{figure}

\paragraph{Beyond-threshold goal definition.}\label{sec:bey}

Instead of defining a goal as \emph{belonging to a quantized bin}, we define it as \emph{exceeding a quality threshold}. Let $\tau_1 < \tau_2 < \cdots < \tau_K$ denote ordered thresholds, and let
$q_i = Q(r_i)$
be the quantized level of sample $i$. Under this beyond-threshold formulation, a trajectory with level $q_i$ is a successful demonstration \emph{not only for its own level, but for every threshold no higher than its outcome}. In other words, sample $i$ provides positive supervision for all goals
\[
g_k^{\mathrm{bey}}:\ \ r \ge \tau_{k}, \quad \forall k \le q_i.
\]
This yields an expanded training set
\begin{equation}
\widetilde{\mathcal{D}}_{\mathrm{bey}}
=
\bigcup_{i=1}^{N}
\left\{
(x_i, s_i, g_k^{\mathrm{bey}})
\;:\;
1 \le k \le {q_i},\; k \in \mathcal{H}
\right\},
\end{equation}
where $\mathcal{H}\subseteq \{1,\ldots,K\}$ denotes the set of thresholds retained for training. 

We then optimize the same goal-conditioned next-token prediction objective:
\begin{equation}
\mathcal{L}_{\mathrm{bey}}(\theta)
=
-\sum_{(x_i,y_i,g)\in \widetilde{\mathcal{D}}_{\mathrm{bey}}}
\sum_{t=1}^{T_i}
\log p_\theta\!\left(y_{i,t}\mid x_i, g, y_{i,<t}\right).
\end{equation}

This reformulation changes the learning signal qualitatively. The model no longer learns merely what responses inside a particular reward interval look like; instead, it learns what kinds of responses are sufficient to satisfy progressively stronger targets. A trajectory with outcome level $R_3$ is therefore not only an example of ``being in bin $R_3$,'' but also a valid demonstration for achieving all the goals such as ``at least $\tau_1$,'' ``at least $\tau_2$,'' and ``at least $\tau_3$.'' This induces an ordered supervision structure across quality levels, which teaches the model how stronger outcomes subsume given threshold goals.


In this sense, the learning objective is no longer to imitate the average behaviors within a fixed bin; rather, it is to pursue consistently directional progression of quality. By associating each response with the thresholds it satisfies, this construction provides supervision for this shared goal-reaching pattern across progressively stronger targets, e.g., from average to better, and from better to even better. This is especially natural for LLMs, because pretrained LLMs already possess substantial pretrained knowledge and semantic capabilities.
By expressing these targets as natural-language instructions and conditioning training on these goals, we can help adapt the LLM’s pretrained capabilities to threshold-reaching behavior in the downstream task.

Although the training responses in the highest-threshold bin remain unchanged, the model learns under different conditioning and from supervision across thresholds through shared parameters. This can change its behavior on unseen inference cases and enable generalization beyond the average quality of the selected demonstrations, thereby mitigating the bounded-learning effect.  See more discussions on this in Appendix \ref{sec:patterns}.

\paragraph{Natural-language goal representation.}
As discussed above, we further replace symbolic goal tokens with natural-language goal descriptions tailored to LLMs. Specifically, for each threshold $\tau_k$, we construct a short instruction that explicitly specifies both the target and the semantics of the corresponding metric. For example, if the task is fine-tuning LLMs for non-toxic generation, the goal can be written as:
\emph{
``Generate a response with a non-toxicity score greater than $0.85$. Non-toxicity scores range from $0$ to $1$, where larger values indicate less toxic content.''
}

Natural-language goals are much more interpretable than abstract reward tokens and allow the LLM to exploit its pretrained instruction-following, semantic understanding, world knowledge, and extrapolation capabilities. This is particularly important under our beyond-threshold goal definition. In this setting, the model should not merely memorize reward buckets; rather, it should understand the meaning of an objective such as achieving performance \emph{above} a threshold, and use that understanding to generalize beyond the exact training patterns. In other words, once the goal is expressed as an explicit instruction, the LLM can naturally understand and learn the ordered relationships among thresholds, and thus adapt the progression patterns learned from training to the inference stage.


\paragraph{Goal filtering for train-test alignment.}\label{sec:filter}
A practical issue in the above data construction is that most trajectories naturally satisfy the weakest thresholds. If all thresholds are kept, the training set will include many samples for goals corresponding to poor outcomes, even though inference always targets desirable high goals. To better align training with test-time use, we retain only thresholds {above the median} performance of the training set, i.e., we choose $\mathcal{H}$ to include only sufficiently good goals. This focuses learning on the transition from {good} to {better}, rather than from {bad} to {average}, and empirically leads to better downstream behavior. See Section \ref{sec:ablation} and \ref{app:goal_filter} for more discussions.

 Overall, we refer to the threshold-based formulation with symbolic goals as \textbf{GCSL-bey}, and to the version further using natural-language goals as \textbf{GCSL-bey-NL}.


\section{Experiments}\label{sec:exp}

We evaluate our approach on three tasks: non-toxic generation, code generation, and LLMs for recommendation. We compare four variants of GCSL for LLM fine-tuning: \textbf{GCSL} denotes the classic offline GCSL formulation, which retains both the original goal definition (i.e., reaching a quantized target value) and the original goal representation using special tokens; \textbf{GCSL-NL} keeps the same formulation but represents goals in natural language (e.g., \emph{``Generate a response with a non-toxicity score between 0.8 and 0.9''}); as well as our proposed \textbf{GCSL-bey} and \textbf{GCSL-bey-NL}.


Across all experiments, we use Qwen3-4B-Instruct as the default base LLM and adopt LoRA for parameter-efficient fine-tuning. Following Quark, we quantize the scores into $K=5$ bins as a default choice. All experiments are conducted with three different seeds, where the average performance and standard deviation are reported. See Appendix \ref{sec:exp_detail} and \ref{app:additional} for more experimental details and ablations.

\subsection{Non-toxic Generation}

\textbf{Task and Dataset.}  
The task is to fine-tune the LLM to generate less toxic responses. We conduct experiments on the REALTOXICITYPROMPTS dataset \citep{gehman2020realtoxicityprompts}, which contains 100K prompts designed to elicit toxic generations. Following prior work \citep{quark, liu2021dexperts}, we use 85K prompts from the training set. For evaluation, we use the same 5K non-toxic test prompts as in~\citep{quark, liu2021dexperts}, and generate responses with nucleus sampling ($p=0.9$). To construct the offline training set, we first use the base LLM (Qwen3-4B-Instruct-2507) to generate a response for each training prompt. We then follow Quark and use the Perspective API \citep{lees2022new} to evaluate each generated response, assigning scores from 0 (non-toxic) to 1 (toxic). 

\textbf{Baselines and Evaluation Metrics.}  

We compare against several representative offline baselines for LLM alignment and/or toxicity reduction. To implement \textbf{DPO} \citep{rafailov2023direct}, we first instruct the base LLM to generate five different answers for each prompt in the training set, and then create the contrastive pairs by selecting the answer pairs with the largest toxicity score differences. \textbf{KTO} \citep{ethayarajh2024kto} directly aligns LLMs by maximizing utility using unpaired binary feedback.
\textbf{IW-SFT} \citep{qin2025supervised}, a reward-weighted SFT variant, introduces importance-sampling weights to dynamic sample losses.
\textbf{LLMEraser} \citep{ding2024unified} is a recent efficient fine-tuning framework for unlearning undesirable characteristics in LLMs. Also, we include a standard \textbf{SFT} baseline, which fine-tunes the model on the top 25\% of training samples with the highest non-toxicity scores.

Following prior work \citep{quark}, we report \emph{maximum toxicity} as the average of the maximum toxicity score over five generations per prompt, and \emph{toxic probability} as the empirical probability that at least one of the five generations is toxic (toxicity score > 0.5). Both are measured by Perspective API. We also report \emph{perplexity} of the generated outputs under the base model as a proxy for language quality and for the extent to which the fine-tuned model deviates from the original model.

\begin{table}[t]
\centering
\caption{Comparison of different methods for non-toxic generation task.}
\label{tab:toxicity}
\resizebox{\linewidth}{!}{%
\begin{tabular}{c|ccccc|ccc|c}
\toprule
 & SFT & IW-SFT & DPO & KTO & LLMEraser & GCSL & GCSL-NL & GCSL-bey & GCSL-bey-NL \\
\midrule
Avg.max.\,$\downarrow$   & 0.139\tiny{$\pm$0.003} & 0.131\tiny{$\pm$0.004} & 0.142\tiny{$\pm$0.006} & 0.147\tiny{$\pm$0.007} & 0.131\tiny{$\pm$0.005} & 0.134\tiny{$\pm$0.003} & 0.129\tiny{$\pm$0.004} & 0.125{\tiny$\pm$0.002} & \textbf{0.115}\tiny{$\pm$0.004} \\
Prob.\,$\downarrow$      & 0.032\tiny{$\pm$0.001} & 0.026\tiny{$\pm$0.006} & 0.036\tiny{$\pm$0.007} & 0.038\tiny{$\pm$0.005} & 0.028\tiny{$\pm$0.004} & 0.027\tiny{$\pm$0.002} & 0.027\tiny{$\pm$0.001} & 0.025\tiny{$\pm$0.002} & \textbf{0.019}\tiny{$\pm$0.003} \\
Perplexity\,$\downarrow$  & 59.43\tiny{$\pm$0.31} & 59.07\tiny{$\pm$0.25} & 62.11\tiny{$\pm$0.43} & 61.70\tiny{$\pm$0.56} & 57.69\tiny{$\pm$0.49} & 55.09\tiny{$\pm$0.33} & 58.67\tiny{$\pm$0.28} & \textbf{54.02}\tiny{$\pm$0.30} & 58.27\tiny{$\pm$0.21} \\
\bottomrule
\end{tabular}%
}
\vspace{-3mm}
\end{table}

\textbf{Results.}  
Table~\ref{tab:toxicity} presents the results. Overall, {GCSL-bey-NL} achieves the best performance among all offline methods in reducing toxic generations while maintaining competitive language quality. 
Comparing {GCSL-bey} and {GCSL-NL} against {GCSL} shows that both the beyond-threshold goal definition and natural-language goal representation lead to clear improvements. Moreover, {GCSL-bey-NL} delivers the strongest gains, suggesting that these two design choices are complementary: the new goal definition provides a more effective optimization target, while natural-language goals allow the LLM to better leverage its semantic understanding of the intended objective.


\subsection{Code Generation}

\textbf{Task and Dataset.}  
The target is to generate both \textbf{correct and efficient} code for given programming problems. We conduct experiments on the Mercury benchmark \citep{du2024mercury}, which contains 1,889 Python programming tasks spanning three difficulty levels. A key feature of Mercury is its \emph{Beyond Score}, a metric for evaluating code efficiency by comparing a candidate solution against a set of reference implementations with different efficiency levels for the same task. 

We randomly sample 20K task-code pairs from Mercury’s training split, using Beyond Score as the reward signal. For evaluation, we use the given test split of Mercury. Following the benchmark protocol, generated code is first executed in an isolated sandbox environment, then its Beyond Score is computed by comparing its execution behavior and efficiency against the reference solutions.

\textbf{Baselines and Evaluation Metrics.}  
Similarly, we include \textbf{DPO}, \textbf{KTO}, and \textbf{IW-SFT} for comparison. To construct the pairwise preference data for DPO, we create five contrastive pairs for each training task by selecting the top five pairs of solutions with the largest runtime differences, following the Mercury paper \citep{du2024mercury}. We further compare with two recent variants of DPO: \textbf{SimPO} \citep{meng2024simpo} removes DPO’s dependence on a reference model; \textbf{AlphaDPO} \citep{wu2025alphadpo} introduces adaptive preference optimization to mitigate the static-reference issue. We also compare SFT on training samples with the highest 25\% Beyond Scores.

Following Mercury, we report \emph{Beyond Score} as the primary metric. As illustrated above, this metric evaluates code efficiency relative to the distribution of valid solutions in the benchmark. Note that Beyond Score also reflects functional correctness, since solutions that fail the tests are directly assigned a score of 0. In addition, we report \emph{Pass Rate} to directly measure the functional correctness.



\begin{table}[t]
\centering
\caption{Comparison of different methods for code generation task.}
\label{tab:mercury}
\small
\resizebox{\linewidth}{!}{
\begin{tabular}{l|cccc|cccc}
\toprule
\multirow{2}{*}{Model} & \multicolumn{4}{c|}{Pass Rate ($\uparrow$)} & \multicolumn{4}{c}{Beyond Score ($\uparrow$)} \\
\cmidrule(lr){2-5} \cmidrule(lr){6-9}
& Easy & Medium & Hard & Overall & Easy & Medium & Hard & Overall \\
\midrule
SFT & 0.888{\tiny$\pm0.011$} & 0.784{\tiny$\pm0.006$} & 0.505{\tiny$\pm0.007$} & 0.723{\tiny$\pm0.007$} & 0.673{\tiny$\pm0.013$} & 0.547{\tiny$\pm0.010$} & 0.452{\tiny$\pm0.006$} & 0.563{\tiny$\pm0.009$} \\
IW-SFT & 0.886{\tiny$\pm0.007$} & 0.782{\tiny$\pm0.005$} & 0.538{\tiny$\pm0.004$} & 0.734{\tiny$\pm0.005$} & 0.752{\tiny$\pm0.010$} & 0.580{\tiny$\pm0.004$} & 0.450{\tiny$\pm0.010$} & 0.602{\tiny$\pm0.006$} \\
DPO & 0.878{\tiny$\pm0.018$} & 0.687{\tiny$\pm0.010$} & 0.445{\tiny$\pm0.009$} & 0.675{\tiny$\pm0.011$} & 0.637{\tiny$\pm0.018$} & 0.450{\tiny$\pm0.014$} & 0.374{\tiny$\pm0.019$} & 0.498{\tiny$\pm0.017$} \\
KTO & 0.882{\tiny$\pm0.016$} & 0.681{\tiny$\pm0.011$} & 0.470{\tiny$\pm0.012$} & 0.685{\tiny$\pm0.013$} & 0.657{\tiny$\pm0.020$} & 0.443{\tiny$\pm0.009$} & 0.394{\tiny$\pm0.014$} & 0.512{\tiny$\pm0.014$} \\
SimPO & 0.880{\tiny$\pm0.009$} & 0.690{\tiny$\pm0.013$} & 0.503{\tiny$\pm0.007$} & 0.698{\tiny$\pm0.010$} & 0.669{\tiny$\pm0.012$} & 0.523{\tiny$\pm0.010$} & 0.392{\tiny$\pm0.008$} & 0.534{\tiny$\pm0.011$} \\
AlphaDPO & 0.884{\tiny$\pm0.010$} & 0.694{\tiny$\pm0.008$} & 0.528{\tiny$\pm0.014$} & 0.710{\tiny$\pm0.010$} & 0.671{\tiny$\pm0.016$} & 0.545{\tiny$\pm0.010$} & 0.523{\tiny$\pm0.017$} & 0.588{\tiny$\pm0.015$} \\
\midrule
GCSL & 0.894{\tiny$\pm0.012$} & 0.692{\tiny$\pm0.008$} & 0.540{\tiny$\pm0.009$} & 0.718{\tiny$\pm0.009$} & 0.893{\tiny$\pm0.013$} & 0.541{\tiny$\pm0.007$} & 0.448{\tiny$\pm0.009$} & 0.650{\tiny$\pm0.009$} \\
GCSL-NL & 0.920{\tiny$\pm0.014$} & 0.700{\tiny$\pm0.011$} & 0.542{\tiny$\pm0.008$} & 0.731{\tiny$\pm0.011$} & 0.918{\tiny$\pm0.016$} & 0.543{\tiny$\pm0.012$} & 0.452{\tiny$\pm0.010$} & 0.662{\tiny$\pm0.012$} \\
GCSL-bey & \textbf{0.959}{\tiny$\pm0.011$} & 0.780{\tiny$\pm0.009$} & 0.540{\tiny$\pm0.007$} & 0.764{\tiny$\pm0.008$} & 0.920{\tiny$\pm0.014$} & 0.612{\tiny$\pm0.008$} & 0.446{\tiny$\pm0.009$} & 0.676{\tiny$\pm0.011$} \\
\midrule
GCSL-bey-NL & 0.924{\tiny$\pm0.014$} & \textbf{0.828}{\tiny$\pm0.006$} & \textbf{0.584}{\tiny$\pm0.012$} & \textbf{0.777}{\tiny$\pm0.009$} & \textbf{0.922}{\tiny$\pm0.012$} & \textbf{0.614}{\tiny$\pm0.008$} & \textbf{0.545}{\tiny$\pm0.012$} & \textbf{0.714}{\tiny$\pm0.010$} \\
\bottomrule
\end{tabular}
}
\vspace{-2mm}
\end{table}

\textbf{Results.} Table~\ref{tab:mercury} summarizes the results. The overall conclusion is similar: {GCSL-bey-NL} achieves the strongest performance, and its advantage is particularly pronounced on more difficult coding tasks. This is intuitive, since harder problems typically exhibit larger performance gaps among candidate solutions. In such cases, learning from fine-grained quality differences and their ordered relationships by GCSL is more beneficial.


Comparing {GCSL-bey} and {GCSL-NL} with {GCSL} also shows that both the beyond-threshold goal formulation and natural-language goal representation contribute clear gains. And the stronger results of {GCSL-bey-NL} indicate that these two components are complementary, and that combining them yields the most effective formulation.

Notably, GCSL-based methods show an even larger advantage on code generation tasks. We attribute this to the task nature: generating code that is both functionally correct and computationally efficient is a particularly subtle objective. It requires the model not only to resolve the coding problems, but also to extrapolate nuanced efficiency differences among valid solutions. This is precisely where GCSL is especially well suited, as it enables the model to learn from structured, graded supervision over outcome quality rather than from binary filtering or pairwise comparisons alone.

\subsection{LLM for Recommendation}\label{sec:rec}

\textbf{Task and Dataset.}  
The target is to fine-tune an LLM to recommend items to users. Specifically, given a user’s historical interaction sequence, the model is asked to predict the item(s) that best match the user’s current preference. We conduct experiments on the Amazon Reviews~\citep{he2016ups} dataset on the \textit{CDs \& Vinyl} category. Following prior work \citep{bao2024decoding, chen2024softmax, gao2025sprec},  we filter out user interaction sequences with fewer than 10 entries. We then split the processed data into training, validation, and test sets with an 8:1:1 ratio. 

Each interaction record includes a user rating from 1 to 5 for the target item, which we directly use as the reward signal to derive the goals for GCSL. For evaluation, following prior works \citep{gao2025sprec,bao2025bi}, we prompt the LLM to generate a predicted item, compute semantic similarity scores between the generated output and all candidate items using Sentence Transformer \citep{reimers2019sentence}, rank the entire item set accordingly, and then derive the final top-$k$ recommendation list.

\textbf{Baselines and Evaluation Metrics.}  
We compare against representative and recent methods for LLM fine-tuning in recommendation. \textbf{BIGRec} \citep{bao2025bi} is an established instruction-tuning framework for sequential recommendation. \textbf{RosePO} \citep{liao2024rosepo} is a DPO-based preference optimization framework that combines negative sampling and personalized uncertainty modeling to improve fairness, robustness, and bias mitigation. \textbf{SPRec} \citep{gao2025sprec} is a recent method that alternates between SFT and DPO to iteratively improve user preference estimation. We also include a direct \textbf{SFT} baseline trained on sequences with ratings greater than or equal to 3, following the standard positive-label threshold commonly adopted in prior recommendation studies \citep{gao2025sprec, he2020lightgcn} on the same dataset.

We report two standard ranking metrics. \emph{NDCG@k} \citep{kang2018self} evaluates the quality of the top-$k$ recommendation list by accounting for both the position of relevant items and their graded relevance scores (i.e., users' ratings). \emph{ERR@k} \citep{chapelle2009expected} interprets the rating as a probability of user satisfaction and places greater emphasis on ranking highly rated items at the top of the list. 
Notably, both metrics place greater value on recommendation items with higher ratings, which highlights the key advantage of our GCSL strategy to leverage graded supervision.



\textbf{Results.} 
Due to space limitation, we present the results in Table \ref{tab:recommendation} in Appendix \ref{app:recommend}.
The trends are similar: {GCSL-bey-NL} achieves the best performance by a clear margin. Notably, the purely DPO-based method {RosePO} performs worse than direct SFT. We attribute this to the nature of the recommendation task and its data structure. In particular, for each training sequence, we only observe the user’s rating for a single target item. As a result, DPO-based methods must sample other items (usually as the negative ones) from the candidate pool to construct the contrastive pairs. This can be problematic, because the sampled items are not necessarily true negative: they may still be preferred by the user, but simply have not been exposed or recommended to them yet. In contrast, GCSL directly leverages the observed graded feedback without requiring such pair construction, and therefore avoids the noise and ambiguity introduced by sampled negatives.

\subsection{Ablations}\label{sec:ablation}

We conduct additional ablation studies to examine the 
effectiveness and mechanism of our approach.
\textbf{Due to space limitation, tables and figures for the ablation results are presented in Appendix \ref{app:ablation_results}}.

\textbf{Effect of the number of thresholds.}  
As described in Section~\ref{sec:bey}, we obtain thresholds by quantizing the training samples into $K$ bins according to their rewards/performance values. Figure~\ref{fig:thresholds} shows that using a larger number of quantiles generally improves performance, as it enables a more fine-grained partition of the training data. This helps GCSL better capture and extrapolate the progression patterns among quality levels.
However, when the number of quantiles becomes too large, the number of training samples within each bin decreases, especially in the highest bin that is most closely aligned with the inference objective. In this case, learning can become less effective because the model has fewer training examples for high-threshold goals, which can  hurt the final performance. That said, we choose $K=5$ in all the experiments, and the consistently superior performance over compared methods demonstrates the robustness of the threshold choice.

\textbf{Effect of the goal filtering strategy.}  
As described in Section~\ref{sec:filter}, we retain only thresholds {above the median training performance} to align the training and inference objectives. To evaluate this, we compare {GCSL-bey} and {GCSL-bey-NL} with their counterparts that do not use goal filtering. Table~\ref{tab:filter} shows that filtering out low-quality goals is important for maintaining the effectiveness of GCSL, as it ensures the alignment between training-time supervision and the goal pursued at inference time.

\textbf{Comparison with SFT under different data-splitting ratios.}  
Figure~\ref{fig:sft} compares {GCSL-bey-NL} with SFT trained on different positive-data ratios, where SFT selects the top 60\% to 10\% of training samples as positive examples. The results show that SFT can be highly sensitive to this choice. The key trade-off is between the quality of the selected positive subset and its size: stricter filtering yields higher-quality supervision but fewer training samples, while looser filtering increases data quantity at the cost of including lower-quality examples. In contrast, our {GCSL-bey-NL} avoids binarizing the positive samples
and \emph{consistently outperforms SFT across all positive-ratio settings}. This demonstrates the advantage of exploiting the hierarchical structure of graded training data, rather than reducing it to a binary split and treating all positive samples as equally informative.

\textbf{Comparison with online methods.}  
Although our method is designed for purely offline learning settings, we further compare it with online methods like Quark and PPO. As shown in Table~\ref{tab:online}, online methods can achieve stronger performance, which is expected because they are able to iteratively generate new data and optimize beyond the limitations of a fixed offline dataset. However, this advantage comes at a substantial cost: online methods require repeated rollout generation, the training and/or use of an external reward model to evaluate newly generated outputs, and repeated model updates on the expanded data. This can be highly time- and resource-intensive, and in many practical settings an appropriate reward model may not be available. In contrast, {GCSL-bey-NL} is fully offline, offering significant efficiency advantages while consistently outperforming other offline approaches.

\section{Conclusion}

In this work, we present goal-conditioned supervised learning as an offline fine-tuning framework for LLMs. By directly treating feedback signals as explicit goals, our method avoids binarizing training positive samples and paired preference construction. We further introduce a beyond-threshold goal formulation and 
natural-language goal representations to better exploit graded feedback. Extensive experiments demonstrate that our approach consistently outperforms standard offline baselines while preserving the efficiency and data efficiency advantages of supervised learning. We've included more discussions on limitations and future works in Appendix \ref{sec:limit}.

\subsection*{AI use statement}

In this work, we used generative AI tools for implementing methods.
We have not used generative AI tools to help develop theoretical models or conceptual frameworks, formulate mathematical claims, provide critical ingredients for proving mathematical claims, propose or refine hypotheses, design or provide feedback on research  methodology or experiments,  assist with translation, clean and reformat dataset, support qualitative and thematic data analysis, interpret results,
and other tasks to generate synthetic data sets, assist in the writing of proofs, are not applicable to this work.
Additionally, we used generative AI tools for editing a research paper to improve readability. We have reviewed all AI-assisted work. For example, LLM-generated code was carefully verified and tested for correctness. We take responsibility for the final content of this work, including text, claims or artifacts produced with the aid of generative AI.

\subsection*{Reproducibility statement}

We elaborate on the experiment details in the Section \ref{sec:exp} and Appendix \ref{sec:exp_detail}, including datasets, baselines, hyper-parameters, etc. We also release the code and data for reproducibility. Our code and data are accessible at \url{https://anonymous.4open.science/r/GCSL_LLM-F5D5}.

\bibliography{reference}
\bibliographystyle{iclr2027_conference}

\newpage
\appendix

\newpage
\appendix

\section{Results on Recommendation Task}\label{app:recommend}

Table \ref{tab:recommendation} presents the results of GCSL-bey-NL and compared baselines on Amazon Reviews dataset for CDs \& Vinyl recommendation.

\begin{table}[h]
\centering
\caption{Comparison of different methods for recommendation by LLM task.}
\label{tab:recommendation}
\resizebox{\linewidth}{!}{%
\begin{tabular}{c|cccc|ccc|c}
\toprule
 & SFT & BIGRec & RosePO & SPRec & GCSL & GCSL-NL & GCSL-bey & GCSL-bey-NL \\
\midrule
NDCG@10 (\%)$\uparrow$  & 0.67\tiny{$\pm$0.02} & 0.71\tiny{$\pm$0.05} & 0.61\tiny{$\pm$0.03} & 0.79\tiny{$\pm$0.06} & 0.72\tiny{$\pm$0.02} & 0.74\tiny{$\pm$0.04} & 1.07\tiny{$\pm$0.03} & \textbf{1.15}\tiny{$\pm$0.02} \\
ERR@10 (\%)$\uparrow$ & 0.35\tiny{$\pm$0.03} & 0.40\tiny{$\pm$0.04} & 0.29\tiny{$\pm$0.06} & 0.44\tiny{$\pm$0.03} & 0.41\tiny{$\pm$0.02} & 0.43\tiny{$\pm$0.05} & 0.46\tiny{$\pm$0.01} & \textbf{0.54}\tiny{$\pm$0.03} \\
\midrule
NDCG@20 (\%)$\uparrow$  & 0.72\tiny{$\pm$0.01} & 0.76\tiny{$\pm$0.03} & 0.69\tiny{$\pm$0.03} & 0.92\tiny{$\pm$0.05} & 0.72\tiny{$\pm$0.04} & 0.79\tiny{$\pm$0.03} & 1.12\tiny{$\pm$0.02} & \textbf{1.27}\tiny{$\pm$0.04} \\
ERR@20 (\%)$\uparrow$ & 0.37\tiny{$\pm$0.02} & 0.44\tiny{$\pm$0.06} & 0.34\tiny{$\pm$0.04} & 0.47\tiny{$\pm$0.04} & 0.42\tiny{$\pm$0.03} & 0.46\tiny{$\pm$0.05} & 0.50\tiny{$\pm$0.02} & \textbf{0.57}\tiny{$\pm$0.03} \\
\bottomrule
\end{tabular}%
}
\vspace{-2mm}
\end{table}

\section{Ablation Results}\label{app:ablation_results}

\begin{figure}[H]
\centering
\begin{minipage}{0.48\linewidth}
    \centering
    \includegraphics[width=0.95\linewidth]{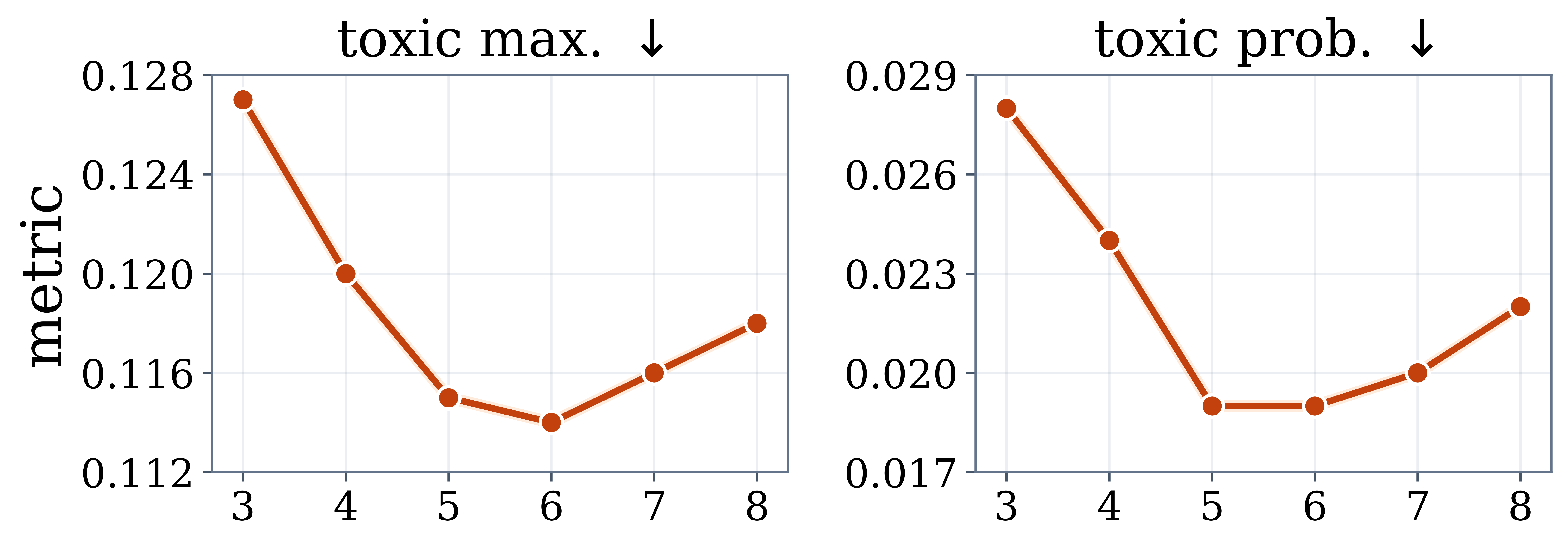}
\end{minipage}
\begin{minipage}{0.47\linewidth}
    \centering
    \includegraphics[width=0.95\linewidth]{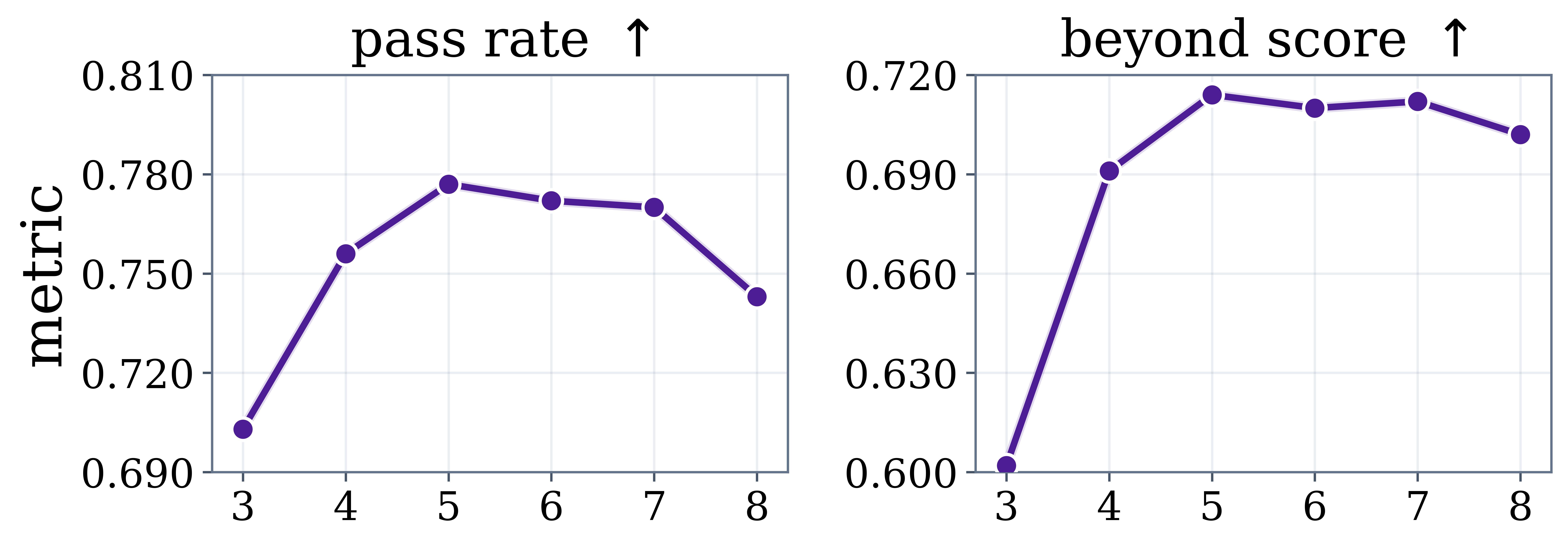}
\end{minipage}
\caption{Performance of GCSL-bey-NL (y-axis) with different number of quantization thresholds (x-axis) on non-toxic generation (left) and code generation (right).}
\label{fig:thresholds}
\vspace{-1mm}
\end{figure}

Figure \ref{fig:thresholds} shows the effect of the threshold numbers on the performance of GCSL-bey-NL.

\begin{figure}[H]
\centering
\begin{minipage}{0.48\linewidth}
    \centering
    \includegraphics[width=0.95\linewidth]{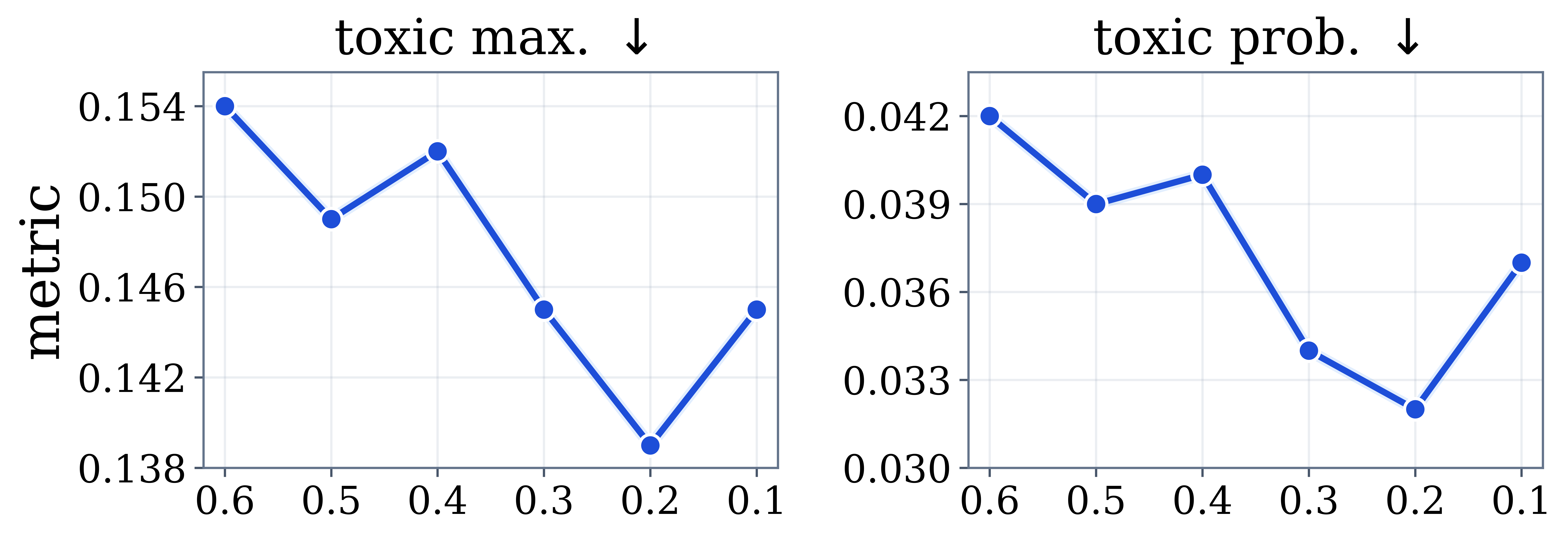}
\end{minipage}
\begin{minipage}{0.48\linewidth}
    \centering
    \includegraphics[width=0.95\linewidth]{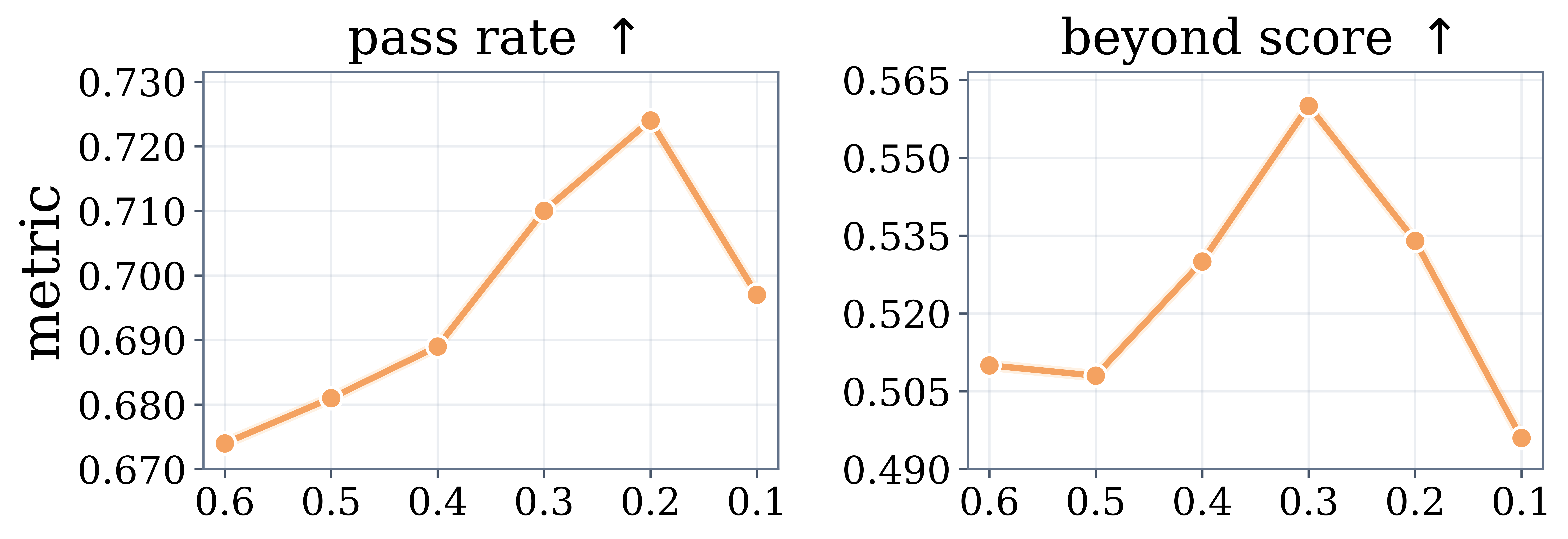}
\end{minipage}
\caption{Performance of SFT (y-axis) with different positive-data ratios (x-axis, i.e., selecting the top 60\% to 10\% of training samples as positive examples) on non-toxic generation (left) and code generation (right).}
\label{fig:sft}
\vspace{-1mm}
\end{figure}

Figure \ref{fig:sft} shows the effect of different positive-data ratios on the performance of SFT. Note that {GCSL-bey-NL achieves 0.115 on toxic max and 0.019 on toxic prob, as well as 0.777 on pass rate and 0.714 on beyond score}. The results show that \textbf{GCSL-bey-NL consistently provides significant performance benefits over SFT across different positive-data ratios}.

Table \ref{tab:filter} compares {GCSL-bey} and {GCSL-bey-NL} with their counterparts that do not use goal filtering. The results demonstrate the effectiveness and necessity of this strategy.

\begin{table}[H]
\centering
\caption{Comparison of GCSL-bey and GCSL-bey-NL with and without goal filtering strategy.}
\label{tab:filter}
\small
\begin{tabular}{c|cc|cc}
\toprule
\multirow{2}{*}{Model} 
& \multicolumn{2}{c|}{RealToxicityPrompts} 
& \multicolumn{2}{c}{Mercury} \\
\cmidrule(lr){2-3} \cmidrule(lr){4-5}
& Avg.max.\,$\downarrow$  & Prob.\,$\downarrow$  
& Pass Rate $\uparrow$ & Beyond Score $\uparrow$ \\
\midrule
GCSL-bey w/o filter     & 0.173\tiny{$\pm$0.008} & 0.049\tiny{$\pm$0.005} & 0.694\tiny{$\pm$0.012}  & 0.634\tiny{$\pm$0.017}   \\
GCSL-bey                & 0.125\tiny{$\pm$0.002} & 0.025\tiny{$\pm$0.002} & 0.764\tiny{$\pm$0.008} & 0.676\tiny{$\pm$0.011}  \\
GCSL-bey-NL w/o filter  & 0.156\tiny{$\pm$0.010}  & 0.035\tiny{$\pm$0.009}  & 0.743\tiny{$\pm$0.015}  & 0.658\tiny{$\pm$0.012}  \\
GCSL-bey-NL             & 0.115\tiny{$\pm$0.004} & 0.019\tiny{$\pm$0.003} & 0.777\tiny{$\pm$0.009} & 0.714\tiny{$\pm$0.010}  \\

\bottomrule
\end{tabular}
\end{table}

Table \ref{tab:online} compares GCSL-bey-NL with online fine-tuning methods such as PPO and Quark. Training time is computed from the start until convergence.
The results show that, although GCSL-bey-NL performs worse than online approaches due to the intrinsic limitations of offline methods trained with fixed offline data, it inherits the significant computational efficiency advantages of standard supervised learning. Specifically, it does not require an external reward model or online labeling/updates, while still consistently outperforming other offline fine-tuning baselines such as SFT.

Notably, the efficiency advantage of offline methods is especially pronounced in the code generation task. This is because online methods must execute each newly generated program in the sandbox environment, compute its performance metrics, and use these scores as rewards for further training. Consequently, the cost and latency of code execution make online approaches particularly expensive and time- and resource-intensive for fine-tuning LLM for code generation. Also, GCSL-bey-NL can approach the performance of online methods on certain metrics (such as the Beyond score with PPO) on the code generation task. We argue that this may be due to the specific curation of the Mercury dataset, with reference solutions spanning the efficiency spectrum, which is favorable to offline approaches compared to a more realistic setting.

\begin{table}[H]
\centering
\caption{Comparison of GCSL-bey-NL and online fine-tuning methods.}
\label{tab:online}
\small
\resizebox{\linewidth}{!}{%
\begin{tabular}{c|ccc|ccc}
\toprule
\multirow{2}{*}{Model} 
& \multicolumn{3}{c|}{RealToxicityPrompts} 
& \multicolumn{3}{c}{Mercury} \\
\cmidrule(lr){2-4} \cmidrule(lr){5-7}
& Avg.max.\,$\downarrow$  & Prob.\,$\downarrow$  & Training Time $\downarrow$ 
& Pass Rate $\uparrow$ & Beyond Score $\uparrow$ & Training Time $\downarrow$ \\
\midrule
PPO           & 0.102\tiny{$\pm$0.008} & 0.013\tiny{$\pm$0.006} & 12.9h & 0.826\tiny{$\pm$0.013} & 0.719\tiny{$\pm$0.015} & 24.5h   \\
Quark         & 0.107\tiny{$\pm$0.005} & 0.015\tiny{$\pm$0.005} & 11.1h & 0.813\tiny{$\pm$0.011} & 0.730\tiny{$\pm$0.015} & 20.8h   \\
SFT           & 0.139\tiny{$\pm$0.003} & 0.032\tiny{$\pm$0.001} & 7.6h & 0.723\tiny{$\pm$0.007} & 0.563\tiny{$\pm$0.009} & 11.6h  \\
GCSL-bey-NL   & 0.115\tiny{$\pm$0.004} & 0.019\tiny{$\pm$0.003} & 8.1h & 0.777\tiny{$\pm$0.009} & 0.714\tiny{$\pm$0.010} & 12.4h   \\

\bottomrule
\end{tabular}
}
\end{table}

\section{Experimental Details}\label{sec:exp_detail}

We conduct all experiments on a dedicated NVIDIA GH200 superchip equipped with an H100 GPU,  with 256 GB of RAM and a 1 TB SSD.
For all models, we use the Adam optimizer and tune the learning rate within the range ([1e-7, 5e-7, 1e-6, 5e-6, 1e-5]). By default, we fine-tune the Qwen3-4B-Instruct-2507 model using LoRA \citep{hu2022lora}, with LoRA rank set to 13, LoRA alpha set to 32, and LoRA dropout set to 0.05. We also set the data type of torch tensor to 16-bit floating point.

When comparing training efficiency, we configure each method with the largest batch size that fits on the H100 GPU. To preserve the language distribution and prevent the model from deviating excessively from the original, we follow Quark \citep{quark} by adding a KL penalty to the loss function for the non-toxic generation task, with the KL coefficient set to 0.05 following Quark. On the recommendation task, we follow the recent LLM-based recommendation setup \citep{gao2025sprec} to further sample 4,096 interactions from the training set, 512 from the validation set, and 1,000 from the test set. For KTO \citep{ethayarajh2024kto} which relies on point-wise binary labels, we use the same 25\% threshold of SFT to split the desirable and undesirable sets.
For all baseline methods, we adhere to the implementation details, optimization configurations, and hyperparameter tuning strategies reported in their original papers. We repeat each experiment three times with different random seeds and report the average and standard deviation of the performance

\section{Additional Experiments and Discussions}\label{app:additional}

\subsection{Comparison with Data Augmentation}

As described in Section~\ref{sec:bey}, our proposed {GCSL-bey} and {GCSL-bey-NL} redefine the GCSL objective as ``exceeding a quality threshold''. Under this formulation, a trajectory with quality level $q_i$ is a successful example not only for its own level, but for every threshold no higher than $q_i$. This expands the training set by associating the same trajectory with multiple valid goals.

One may wonder whether the gains from this design simply come from implicit data augmentation by oversampling. To examine this, we construct a control variant that uses the same data expansion procedure as GCSL-bey but removes goal conditioning. For example, if an offline trajectory $\tau_k$ has quality level $q_3$, we first repeat it three times following the GCSL-bey construction and include all copies as the expanded training data. Then we also apply the same filtering strategy as GCSL-bey, retaining only trajectories with quality level above the median threshold in the original training set. We then perform standard supervised fine-tuning on this expanded dataset without appending any goals. We denote this variant as \textbf{SFT-aug}.

Table~\ref{tab:sft_acc} compares {SFT-aug} with {GCSL-bey} and {GCSL-bey-NL}. The results show that the gains of GCSL-bey are not merely due to repeating high-quality samples more often. Rather, the improvement comes from explicitly conditioning on goals and learning the relationship between trajectories and ordered quality thresholds.

\begin{table}[H]
\centering
\caption{Comparison of SFT-aug with GCSL-bey and GCSL-bey-NL.}
\label{tab:sft_acc}
\small
\begin{tabular}{c|cc|cc}
\toprule
\multirow{2}{*}{Model} 
& \multicolumn{2}{c|}{RealToxicityPrompts} 
& \multicolumn{2}{c}{Mercury} \\
\cmidrule(lr){2-3} \cmidrule(lr){4-5}
& Avg.max.\,$\downarrow$  & Prob.\,$\downarrow$  
& Pass Rate $\uparrow$ & Beyond Score $\uparrow$ \\
\midrule
SFT-aug                 & 0.132\tiny{$\pm$0.007} & 0.029\tiny{$\pm$0.004} & 0.727\tiny{$\pm$0.013} & 0.598\tiny{$\pm$0.017}  \\
GCSL-bey                & 0.125\tiny{$\pm$0.002} & 0.025\tiny{$\pm$0.002} & 0.764\tiny{$\pm$0.008} & 0.676\tiny{$\pm$0.011}  \\
GCSL-bey-NL             & 0.115\tiny{$\pm$0.004} & 0.019\tiny{$\pm$0.003} & 0.777\tiny{$\pm$0.009} & 0.714\tiny{$\pm$0.010}  \\

\bottomrule
\end{tabular}
\end{table}

\subsection{Discussion of Quantization}

As described in Section~\ref{sec:classic_gcsl}, both classic GCSL and our GCSL-bey begin by quantizing feedback signals in the training data into a finite set of goal labels. This design follows prior work on applying GCSL with LLMs or Transformer-based sequential models~\citep{TrajectoryTransformer, quark}. The main motivation is training efficiency: quantization makes the joint distribution over goals and tokenized prompt--response trajectories much denser and therefore easier to learn.

By contrast, if we directly use continuous scalar feedback as goals, the joint space of numerical goals and corresponding trajectories becomes highly sparse, especially in offline settings with limited data. Such sparsity can significantly reduce the learning efficiency of GCSL and significantly harm the final performance, as also noted in prior work~\citep{TrajectoryTransformer, quark}. Nevertheless, we additionally test this issue in our setting by comparing GCSL with and without quantization. Specifically, we compare the classic GCSL adaptation for LLM fine-tuning, which quantizes feedback signals and represents goals with special reward tokens, with a non-quantized variant that maps scalar feedback values to embedding representations via an external reward embedding layer. These embeddings are then provided as additional inputs to the LLM Transformer decoder. We also compare GCSL-NL with its variant without quantization, which converts original scalar feedback values into strings with two decimal places and tokenized by the LLM tokenizer.

Table \ref{tab:quantization} presents the results, demonstrating the necessity and effectiveness of the goal quantization procedure for GCSL in LLM fine-tuning.

\begin{table}[H]
\centering
\caption{Comparison of GCSL and GCSL-NL with their variants without goal quantization.}
\label{tab:quantization}
\small
\begin{tabular}{c|cc|cc}
\toprule
\multirow{2}{*}{Model} 
& \multicolumn{2}{c|}{RealToxicityPrompts} 
& \multicolumn{2}{c}{Mercury} \\
\cmidrule(lr){2-3} \cmidrule(lr){4-5}
& Avg.max.\,$\downarrow$  & Prob.\,$\downarrow$  
& Pass Rate $\uparrow$ & Beyond Score $\uparrow$ \\
\midrule
GCSL w/o quantization   & 0.179\tiny{$\pm$0.007} & 0.049\tiny{$\pm$0.004} & 0.614\tiny{$\pm$0.016} & 0.485\tiny{$\pm$0.011}  \\
GCSL                    & 0.134\tiny{$\pm$0.003} & 0.027\tiny{$\pm$0.002} & 0.718\tiny{$\pm$0.009} & 0.650\tiny{$\pm$0.009}  \\
GCSL-NL w/o quantization  & 0.168\tiny{$\pm$0.005} & 0.046\tiny{$\pm$0.005} & 0.630\tiny{$\pm$0.010} & 0.506\tiny{$\pm$0.013}  \\
GCSL-NL                 & 0.129\tiny{$\pm$0.004} & 0.027\tiny{$\pm$0.001} & 0.731\tiny{$\pm$0.011} & 0.662\tiny{$\pm$0.011}  \\

\bottomrule
\end{tabular}
\end{table}

\subsection{Results with Other Backbone LLM}

In addition to using Qwen3-4B-Instruct-2507 as the base model, we also conducted experiments with a larger backbone, Llama-3.1-8B-Instruct. The corresponding results are reported in Table~\ref{tab:llama} and Table~\ref{tab:llama_2}. The findings consistently demonstrate the robustness of GCSL-bey-NL and its sustained performance advantages over the compared baseline methods. These results further highlight the cross-model transferability of the optimization benefits enabled by the proposed goal-conditional supervised learning framework.

\begin{table}[H]
\centering
\caption{Performance comparison with Llama-3.1-8B-Instruct as backbone on non-toxic generation and code generation tasks.}
\label{tab:llama}
\small
\begin{tabular}{c|cc|cc}
\toprule
\multirow{2}{*}{Model} 
& \multicolumn{2}{c|}{RealToxicityPrompts} 
& \multicolumn{2}{c}{Mercury} \\
\cmidrule(lr){2-3} \cmidrule(lr){4-5}

& Avg.max.\,$\downarrow$  & Prob.\,$\downarrow$
& Pass Rate $\uparrow$ & Beyond Score $\uparrow$ \\
\midrule
SFT
& 0.113\tiny{$\pm$0.002}
& 0.017\tiny{$\pm$0.001}
& 0.635\tiny{$\pm$0.004}
& 0.515\tiny{$\pm$0.003} \\

DPO
& 0.119\tiny{$\pm$0.003}
& 0.021\tiny{$\pm$0.002}
& 0.630\tiny{$\pm$0.008}
& 0.507\tiny{$\pm$0.006} \\

KTO
& 0.117\tiny{$\pm$0.004}
& 0.015\tiny{$\pm$0.002}
& 0.639\tiny{$\pm$0.005}
& 0.521\tiny{$\pm$0.008} \\

\midrule
GCSL
& 0.112\tiny{$\pm$0.003}
& 0.016\tiny{$\pm$0.000}
& 0.629\tiny{$\pm$0.004}
& 0.501\tiny{$\pm$0.005} \\

GCSL-NL
& 0.109\tiny{$\pm$0.002}
& 0.015\tiny{$\pm$0.002}
& 0.634\tiny{$\pm$0.007}
& 0.514\tiny{$\pm$0.007} \\

GCSL-bey
& 0.107\tiny{$\pm$0.001}
& 0.013\tiny{$\pm$0.000}
& 0.662\tiny{$\pm$0.005}
& 0.553\tiny{$\pm$0.004} \\

\midrule
GCSL-bey-NL
& \textbf{0.104}\tiny{$\pm$0.001}
& \textbf{0.010}\tiny{$\pm$0.000}
& \textbf{0.694}\tiny{$\pm$0.006}
& \textbf{0.610}\tiny{$\pm$0.005} \\

\bottomrule
\end{tabular}
\end{table}

\begin{table}[h]
\centering
\caption{Performance comparison with Llama-3.1-8B-Instruct as backbone on recommendation task.}
\label{tab:llama_2}
\resizebox{\linewidth}{!}{%
\begin{tabular}{c|cccc|ccc|c}
\toprule

& SFT & BIGRec & RosePO & SPRec & GCSL & GCSL-NL & GCSL-bey & GCSL-bey-NL \\
\midrule
NDCG@10 (\%)$\uparrow$
& 0.86\tiny{$\pm$0.02}
& 0.92\tiny{$\pm$0.07}
& 0.78\tiny{$\pm$0.03}
& 1.02\tiny{$\pm$0.06}
& 0.93\tiny{$\pm$0.04}
& 0.96\tiny{$\pm$0.08}
& 1.37\tiny{$\pm$0.02}
& \textbf{1.49}\tiny{$\pm$0.05} \\

ERR@10 (\%)$\uparrow$
& 0.45\tiny{$\pm$0.04}
& 0.51\tiny{$\pm$0.06}
& 0.37\tiny{$\pm$0.02}
& 0.57\tiny{$\pm$0.07}
& 0.53\tiny{$\pm$0.03}
& 0.55\tiny{$\pm$0.05}
& 0.60\tiny{$\pm$0.01}
& \textbf{0.70}\tiny{$\pm$0.06} \\

\midrule
NDCG@20 (\%)$\uparrow$
& 0.93\tiny{$\pm$0.03}
& 0.98\tiny{$\pm$0.05}
& 0.89\tiny{$\pm$0.01}
& 1.18\tiny{$\pm$0.08}
& 0.96\tiny{$\pm$0.04}
& 1.02\tiny{$\pm$0.06}
& 1.45\tiny{$\pm$0.02}
& \textbf{1.63}\tiny{$\pm$0.07} \\

ERR@20 (\%)$\uparrow$
& 0.48\tiny{$\pm$0.02}
& 0.57\tiny{$\pm$0.07}
& 0.44\tiny{$\pm$0.04}
& 0.61\tiny{$\pm$0.06}
& 0.54\tiny{$\pm$0.03}
& 0.59\tiny{$\pm$0.08}
& 0.64\tiny{$\pm$0.02}
& \textbf{0.74}\tiny{$\pm$0.05} \\

\bottomrule
\end{tabular}%
}
\vspace{-2mm}
\end{table}

\subsection{Results with Flexible Thresholds}\label{app:flex_thre}

As discussed in Section~\ref{sec:bey_gcsl}, {GCSL-bey-NL} defines goals based on the quantized thresholds derived from the training data, and uses the highest threshold as the inference-time goal for pursuing high-quality outcomes. Since a key motivation of our work is to leverage the flexible semantic understanding and instruction-following ability of LLMs to generalize goal-achieving behavior, we further study performance under more flexible goals that do not explicitly appear in the training set. Specifically, we scale the highest training threshold by a series of factors and define new inference goals using the resulting values, keeping the same decimal precision as the original threshold. For example, if the highest training threshold is 0.80 and the scaling factor is 1.04, the corresponding goal becomes: \emph{``Generate a response with a non-toxicity score greater than 0.83. Non-toxicity scores range from 0 to 1, where larger values indicate less toxic content.''}

\begin{figure}[h]
\centering
\begin{minipage}{0.48\linewidth}
    \centering
    \includegraphics[width=0.95\linewidth]{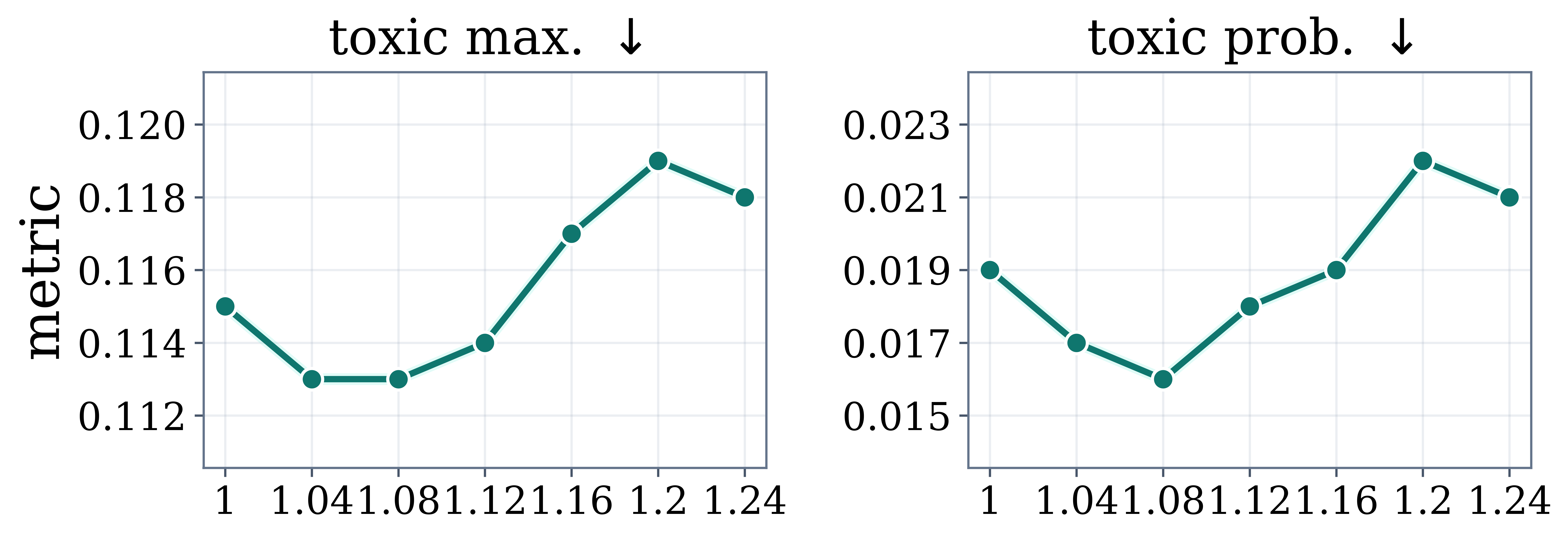}
\end{minipage}
\begin{minipage}{0.48\linewidth}
    \centering
    \includegraphics[width=0.95\linewidth]{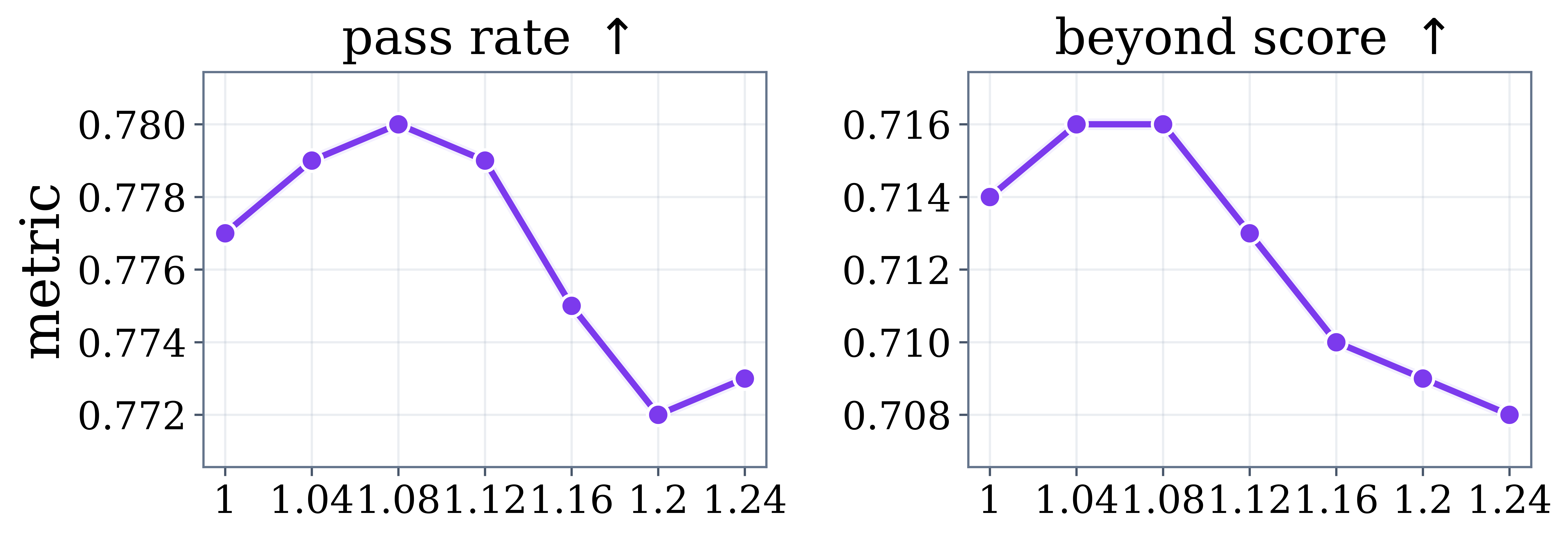}
\end{minipage}
\caption{Performance of GCSL-bey-NL (y-axis) with different scaling factors for inference goal (x-axis) on non-toxic generation (left) and code generation (right).}
\label{fig:scale}
\vspace{-1mm}
\end{figure}

Figure~\ref{fig:scale} shows the results. As the scaling factor increases from 1.04 to 1.24, performance first improves slightly and then declines. This is reasonable: when the inference goal is close to the highest training objective but slightly more ambitious, the model is able to use its semantic knowledge and inherent reasoning ability to adapt its behavior toward a better target, leading to improved performance. However, when the new goal becomes too far from the training distribution, the model may no longer be able to effectively transfer the knowledge learned during post-training to the downstream task, which leads to performance degradation.

Nevertheless, these ablations still perform generally better than most baselines in our main experiments. The results suggest that the model remains reasonably effective even when queried with unseen, interpolated, but reasonable thresholds at inference time. This supports our claim that natural-language goal representations enable the model to leverage its semantic understanding and generalize the goal-achieving behavior learned during training to more flexible inference-time objectives.

\subsection{Discussion of Goal Presentation}

As illustrated in Section \ref{sec:bey_gcsl}, the goals in GCSL-bey and GCSL-bey-NL are represented by special tokens and concise instruction prompts for specifying the task target and the semantics of the corresponding metric, respectively. More specifically, we define the special tokens like [\texttt{\_TREE\_TOKEN\_0}, ..., \texttt{\_TREE\_TOKEN\_4}] and register them in the model’s tokenizer configuration. For natural language goals, we use the following for each experimental task. 

On the non-toxic generation task, the NL goal is like:

\emph{``Generate a response with a non-toxicity score greater than 0.80. Non-toxicity scores range from 0 to 1, with 1 indicating the least toxicity.''}

On the efficient code generation task, the NL goal is like:

\emph{``Follow the instructions to generate a correct Python solution with an Efficiency Score greater than 0.82. Efficiency scores range from 0 to 1, with 1 indicating highest efficiency.''}

On the LLM for recommendation task, the NL goal is like:

\emph{``Generate a recommendation with a user preference rating equal or greater than 4. Ratings range from 1 to 5, with 5 indicating the highest rating.''}

Furthermore, we conduct experiments to examine how sensitive {GCSL-bey-NL} is to the exact wording of the natural-language goals. Specifically, we remove the detailed task instruction and metric explanation from the goal prompt, and retain only the metric name and target threshold. For example, in the non-toxicity generation task and the efficient code generation task, the simplified goals are: \emph{``Generate a response with a non-toxicity score greater than 0.80.''} and \emph{``Generate a Python solution with an Efficiency Score greater than 0.82.''}, respectively. We denote this simplified-language variant as {GCSL-bey-SNL}.

Table~\ref{tab:simple} shows that {GCSL-bey-SNL} still significantly outperforms SFT and GCSL-bey on both tasks, demonstrating that the gains from natural-language goal presentation are robust and can effectively leverage the LLM’s language understanding and generalization capabilities. Nevertheless, the performance gap between {GCSL-bey-SNL} and {GCSL-bey-NL} also highlights the value of presenting goals more clearly, with explicit task context and metric interpretation. This further supports the importance of goal representation design when applying {GCSL-bey} to LLM fine-tuning.

\begin{table}[H]
\centering
\caption{Comparison with GCSL-bey-SNL for simple goal representation.}
\label{tab:simple}
\small
\begin{tabular}{c|cc|cc}
\toprule
\multirow{2}{*}{Model} 
& \multicolumn{2}{c|}{RealToxicityPrompts} 
& \multicolumn{2}{c}{Mercury} \\
\cmidrule(lr){2-3} \cmidrule(lr){4-5}
& Avg.max.\,$\downarrow$  & Prob.\,$\downarrow$  
& Pass Rate $\uparrow$ & Beyond Score $\uparrow$ \\
\midrule
SFT                     & 0.139\tiny{$\pm$0.003} & 0.032\tiny{$\pm$0.001} & 0.723\tiny{$\pm$0.007} & 0.563\tiny{$\pm$0.009}  \\
GCSL-bey                & 0.125\tiny{$\pm$0.002} & 0.025\tiny{$\pm$0.002} & 0.764\tiny{$\pm$0.008} & 0.676\tiny{$\pm$0.011}  \\
\midrule
GCSL-bey-SNL            & {0.119}\tiny{$\pm$0.001} & {0.022}\tiny{$\pm$0.000} & {0.772}\tiny{$\pm$0.004} & {0.703}\tiny{$\pm$0.006}  \\
GCSL-bey-NL             & {0.115}\tiny{$\pm$0.004} & {0.019}\tiny{$\pm$0.003} & {0.777}\tiny{$\pm$0.009} & {0.714}\tiny{$\pm$0.010}  \\

\bottomrule
\end{tabular}
\end{table}

\subsection{More Discussion of Goal Filtering}\label{app:goal_filter}

We've discussed the motivation and reasonability of our goal filtering strategy in Section \ref{sec:filter}, and presented experimental results supporting its effectiveness and necessity in Section \ref{sec:ablation} and Appendix \ref{app:ablation_results}. Here, we conduct a more in-depth discussion of this goal filtering strategy.

Specifically, we focus on the cutoff of our goal filtering strategy. As discussed in Section \ref{sec:filter}, our motivation is to align the training process with the inference target of GCSL-bey-NL, i.e., the focus of learning should be on the transition from good to better, rather than from bad to average. As a result, we filter out goals below the median performance on the training set as a direct and default strategy. However, it would still be meaningful to study this cutoff in a more fine-grained way. To this end, we compare GCSL-bey-NL by setting the cutoff of the goal filtering strategy among ratios of 0\%, 25\%, 50\%, and 75\%. For example, if the ratio is 25\%, we discard the goals that underperform the lowest 25\% of the training samples when training our method.

\begin{figure}[H]
\centering
\begin{minipage}{0.48\linewidth}
    \centering
    \includegraphics[width=0.95\linewidth]{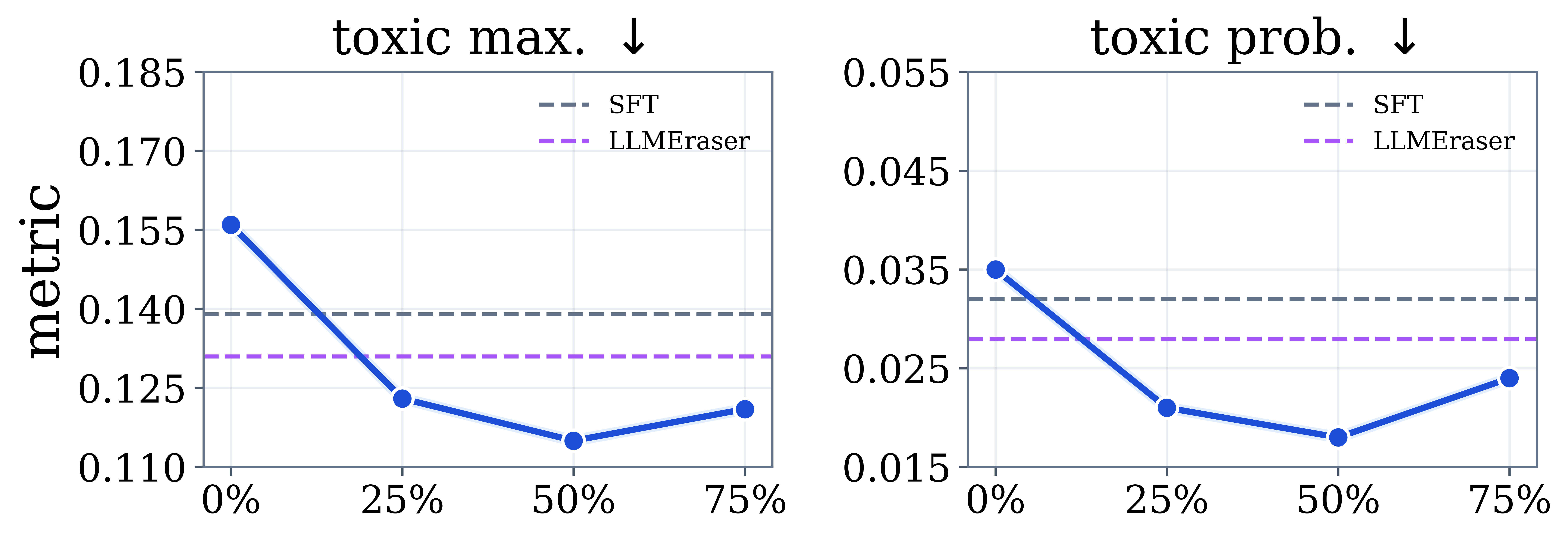}
\end{minipage}
\begin{minipage}{0.48\linewidth}
    \centering
    \includegraphics[width=0.95\linewidth]{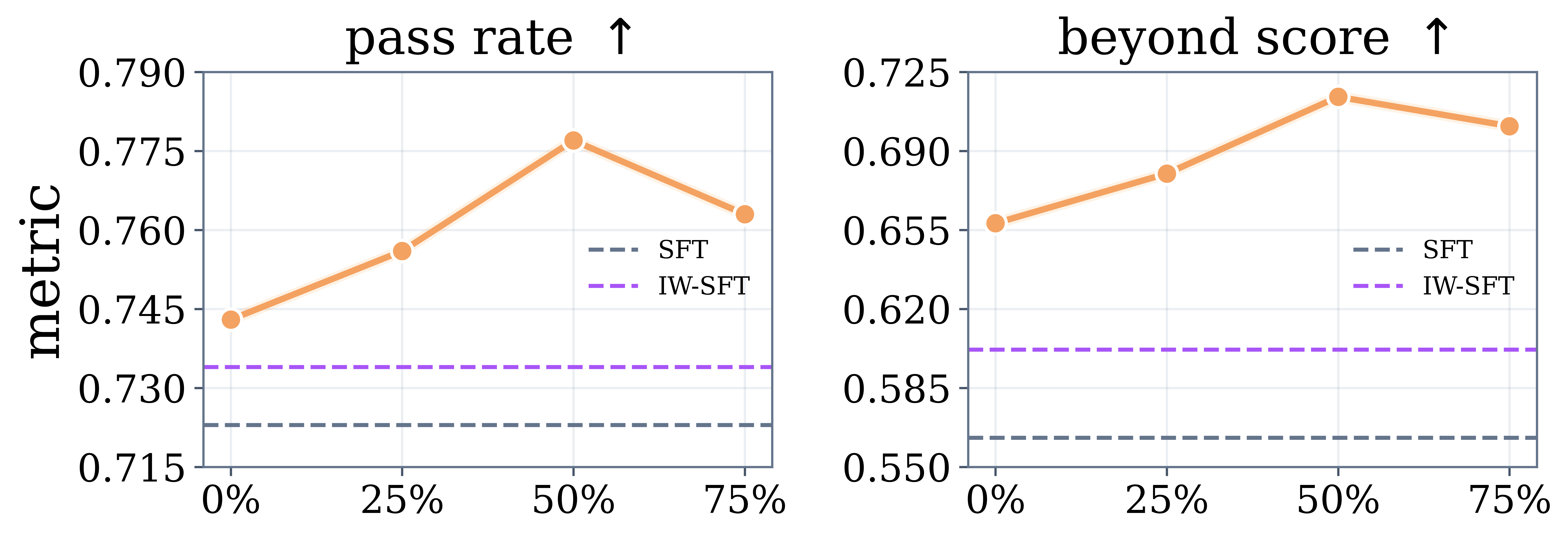}
\end{minipage}
\caption{Performance of GCSL-bey-NL (y-axis) with different goal-filtering ratios (x-axis, i.e., filtering the bottom 75\% to 25\% of training samples as positive examples) on non-toxic generation (left) and code generation (right).}
\label{fig:goal_filter}
\vspace{-1mm}
\end{figure}

Figure~\ref{fig:goal_filter} shows the results. The trends align with our previous assumption: filtering out low-quality goals is important for improving the performance of GCSL, as it ensures alignment between training-time supervision and the goal pursued at inference time. Also, setting the threshold at 50\% is a generally robust choice. If we set the filtering threshold too high, although we retain more high-goal-oriented data for training, the size of the training data with diverse goals, along with the demonstrations for beyond-threshold goal-reaching patterns, also decreases significantly, which can instead harm the performance of GCSL-bey-NL.

We also compare SFT and the best-performing baselines on each dataset (LLMEraser and IW-SFT, respectively). As shown in the figure, GCSL-bey-NL outperforms all baselines except at 0\% (no goal filtering) on non-toxic generation, and outperforms all baselines even without goal filtering on code generation. These results demonstrate the robustness of GCSL-bey-NL's design: we only need to filter out the worst goals for training, and GCSL-bey-NL can already achieve strong performance with negligible effort on hyperparameter choice or tuning.

\subsection{Discussion on Learning Patterns}\label{sec:patterns}

We'd like to address that we do not attribute the superior performance of {GCSL-bey-NL} to an ``unbounded'' set of training demonstrations for the maximum goal used at inference. Rather, as discussed in Section~\ref{sec:bey_gcsl}, {GCSL-bey-NL} is designed to learn a general goal-achieving behavior from training samples by conditioning on the objective of exceeding a given threshold, which naturally encourages improvement toward better outcomes. At inference time, the fine-tuned LLM is expected to extrapolate this learned goal-achieving progression pattern beyond the exact training goals, leveraging the strong extrapolation, parameter sharing, and natural-language understanding capabilities of LLMs. In this way, the model can transfer the learned goal-achieving patterns to pursue stronger performance at inference time.


This kind of generalization and extrapolation is consistent with the capabilities demonstrated by recent strong LLMs. For example, one-shot and few-shot prompting have shown that LLMs can learn effectively from only a small number of examples in context \citep{agarwal2024many, xu2024does}. Although such examples provide only a limited set of task instances, the model can still leverage its internal knowledge and reasoning ability to generalize to related unseen cases and achieve strong performance. 

Our approach goes beyond standard in-context learning. We explicitly construct training samples associated with our defined goals and fine-tune the LLM to learn how to achieve them. In this way, the model can have a deeper understanding of general goal-achieving patterns for pursuing progressive outcomes beyond given performance thresholds on specific downstream task, and then extrapolate these learned patterns at inference time. As a result, it has the ability to achieve performance beyond the average behavior of the highest training bin, even when such superior patterns may not explicitly present in the training data.

\subsection{Expanded Related Work}\label{sec:exp_rel}

Some recent works have also explored offline optimization of LLMs from reward-like supervision, but these formulations remain meaningfully different from ours. Swift \citep{mukherjee2025offline} is related in spirit of utilizing ordered rewards, but its formulation and proposed methodology are specialized to conversations with fixed turns. As a result, it targets a much narrower interaction setting than our work, which studies general offline LLM fine-tuning over open-ended prompt--response data. Our method instead focuses on a broader next-token generation setting, where graded feedback is incorporated directly through goal-conditioned supervised learning without assuming a fixed conversational horizon. RRHF \citep{yuan2023rrhf} also explores efficient LLM alignment by combining a response-ranking objective with supervised fine-tuning. However, its ranking objective requires multiple responses to the same prompt and their relative quality assessments to construct comparisons. In contrast, our method directly uses the feedback associated with each individual prompt–response example as goal supervision, allowing training even when only one observed response is available per prompt.

CA \citep{hu2023aligning} is also related in that it replaces online PPO-style alignment with offline objectives. However, it still requires training a separate human preference model, and its conditional alignment strategy leaves the inference-time reward choice unclear and not principled: for a given downstream task, it is unclear what reward value should be specified at deployment. By contrast, our framework defines the inference objective directly through explicit goals directly derived from observed feedback, and our beyond-threshold formulation provides a clearer and more natural target for controlled generation. Some recent preference-based methods like KTO \citep{ethayarajh2024kto} further explore the use of unpaired preferences for LLM fine-tuning. However, they still strictly rely on binary labels. In contrast, our proposed GCSL-bey-NL can directly leverage fine-grained graded feedback, avoiding the coarse treatment of feedback signals.

Our work is also related in motivation to ordinal labeling \citep{diaz2019soft}, since both seek to preserve the ordered structure in graded supervision rather than collapsing it into binary labels. However, ordinal-labeling methods are primarily developed for ordered multi-class classification, where the output is a discrete label or label distribution. In contrast, our setting is autoregressive next-token generation for LLM fine-tuning, where graded feedback must be incorporated into the conditioning and learning objective of a generative model. Therefore, although ordinal labeling is conceptually relevant, it is not directly comparable to our setting.

Last, while our approach is directly motivated by extending GCSL to LLM fine-tuning, related ideas for learning from graded reward signals were also explored in earlier command-conditioned RL research, particularly Upside-Down RL \citep{schmidhuber2019reinforcement, ashley2022learning}. Specifically, its relational-command formulation allows observed trajectories to supervise commands requesting returns above a given threshold, sharing the underlying principle of our beyond-threshold relabeling. Our work develops similar principle in conjunction with natural-language goal conditioning for fully offline LLM fine-tuning, using semantically meaningful instructions and goal filtering to support learning from graded feedback. Through controlled experiments across three tasks, we demonstrate the complementary benefits of these components and identify design choices critical to their effectiveness in this setting.




\subsection{Discussion of the Bounded-Learning Effect}

In Sections \ref{sec:intro} and \ref{sec:def_GCSL}, we discussed the bounded-learning effect of SFT and conventional GCSL, where learning is implicitly biased toward the average quality of the selected training subset. In this section, we further examine this effect through their optimization objectives and clarify its implications for our proposed approach.

The underlying intuition follows directly from how these methods use demonstrations. SFT maximizes the likelihood of responses in a selected positive subset, while conventional GCSL maximizes their likelihood conditioned on the corresponding reward bin. Within each subset, responses receive the same training weight regardless of their quality differences. Consequently, both objectives encourage matching the observed response distribution, without explicitly favoring its higher-quality outcomes. This interpretation is consistent with prior analyses of return-conditioned supervised learning \citep{brandfonbrener2022does} and SFT on curated data \citep{qin2025supervised}, which connect their supervised objectives to matching the conditional behavior distribution and the selected demonstration distribution, respectively.

More formally, let \(\mathcal D_g\) denote the selected subset for SFT or the subset associated with goal \(g\) for conventional GCSL. Let \(d_g(x)\) and \(q_g(s\mid x)\) denote its empirical prompt distribution and conditional response distribution, respectively. The normalized supervised objective can be written as

$$
\mathcal L_g(\theta)
=
-\mathbb E_{x\sim d_g,\,s\sim q_g}
\log p_\theta(s\mid x,g)
=
C_g+
\underbrace{\mathbb E_{x\sim d_g}
D_{\mathrm{KL}}\!\left(
q_g(\cdot\mid x)\,\|\,p_\theta(\cdot\mid x,g)
\right)}_{\epsilon_g(\theta)},
$$

where \(C_g\) is independent of \(\theta\), and goal conditioning is omitted for SFT. In the realizable case, the global optimum therefore matches \(q_g\), and its expected quality under the same prompt distribution equals the subset’s average quality.

This conclusion also admits an approximate characterization. For a fixed quality function \(r(x,s)\in[r_{\min},r_{\max}]\), define \(\bar r_g=\mathbb E_{d_g,q_g}[r(x,s)]\) and \(J_g(\theta)=\mathbb E_{d_g, p_\theta}[r(x,s)]\). Pinsker’s inequality and Jensen’s inequality yield

$$
\left|J_g(\theta)-\bar r_g\right|
\le
(r_{\max}-r_{\min})
\sqrt{\frac{\epsilon_g(\theta)}{2}}.
$$

Thus, increasingly accurate distribution matching brings expected quality toward the subset mean, formally explaining the bounded-learning tendency induced by these objectives.

The above analysis characterizes distribution matching under the empirical prompt distribution and also applies to GCSL-bey at a fixed threshold. It does not establish an upper bound on performance for unseen prompts. Our mitigation claim concerns how supervision across thresholds, particularly when expressed through natural-language goals, adapts a pretrained LLM’s existing capabilities to the task. Such adaptation can change generalization behavior even when the highest-threshold demonstrations remain identical. The analysis characterizes the imitation tendency, while our experiments evaluate the effectiveness of the proposed goal-conditioned adaptation. 



\subsection{Discussion on Social Impact}

Our work aims to provide a more efficient and practical framework for LLM fine-tuning in settings with graded feedback, which may benefit applications such as safer text generation, better code assistance, and more accurate recommendation. By avoiding reward-model-based online optimization, the proposed method can lower the computational and engineering cost of alignment, making such techniques more accessible.

At the same time, the method inherits the limitations of the feedback it learns from. If the feedback signals are biased, incomplete, or misaligned with user welfare, the fine-tuned model may reproduce or amplify these issues. This is especially important in application domains such as recommendation, where optimization may unintentionally reinforce popularity bias or narrow user exposure. Therefore, careful dataset design, feedback auditing, and task-specific safety evaluation remain important when deploying our approach in practice.

\section{Limitations and Future Works}\label{sec:limit}

Although {GCSL-bey-NL} demonstrates strong effectiveness and efficiency for offline LLM fine-tuning, we identify some potential limitations and corresponding future works. 

First, our method is designed and evaluated in a purely offline setting. While this is an important advantage in scenarios where online rollouts or reward models are costly or unavailable, {GCSL-bey-NL} can in principle also be extended to online learning in a manner similar to Quark. It would therefore be interesting to study how our beyond-threshold goal formulation and natural-language goal representation behave when combined with iterative data collection and online policy improvement.

Second, our current framework still relies on quantization to construct a finite set of goal levels. As discussed earlier, this design is helpful and necessary for training efficiency and data density, but the choice of quantization granularity still requires specific parameter tuning. Future work could explore more adaptive goal discretization strategies, or hybrid formulations that better bridge discrete goal labels and continuous feedback signals.

Third, in this paper we focus on scalar or categorical feedback, which already covers many practical fine-tuning scenarios and allows us to study the proposed framework in a clean and controlled setting. However, real-world applications may also involve richer supervision signals, such as multi-aspect judgments or more dynamic forms of user preference. Specifically, our framework is expected to be easily extended to multi-objective LLM alignment scenarios, where the only adaptation needed is to extend the goal definition and expression from a single objective to multiple objectives. For example, one could define a goal as \emph{
``Generate a response with a non-toxicity score greater than $0.85$ and perplexity score greater than $58.2$.''} To keep the scope of this paper focused, we restrict the current discussion to the single-objective setting, and leave the extension of GCSL to these richer settings for future work.

Last, future research can also extend or apply our method and experiments to broader areas. For example, although our experiments in Appendix \ref{app:flex_thre} show that one cannot indefinitely increase the threshold used at inference time without degrading performance, exploring the possibility of assigning more ambitious inference goals is still an interesting topic to explore. For instance, adapting GCSL-bey-NL to an online setting with iterative data collection and online policy improvement would likely increase the magnitude of effective inference goals over time. Furthermore, we conduct our experiments using standard LLM alignment/fine-tuning benchmarks on non-toxic generation, code generation, and commercial item recommendation. Nevertheless, exploring the application of GCSL-bey-NL in large-scale industrial settings with tremendous amounts of data would be a valuable future direction, especially for industrial researchers with ample resources.




\end{document}